\documentclass[journal]{IEEEtran}

\usepackage{graphicx}  
\usepackage{epstopdf}  
\usepackage{multirow}
\usepackage{array}  
\usepackage{color}
\usepackage{colortbl}
\usepackage{booktabs}
\usepackage{amsmath}
\usepackage{mathrsfs}  
\usepackage{amsfonts}  
\usepackage[switch]{lineno}
\usepackage{algorithm}
\usepackage{algpseudocode}
\usepackage{amsmath}
\usepackage{booktabs}
\usepackage{threeparttable}
\usepackage{url}
\usepackage{makecell, verbatim, sidecap}

\begin{document}
	
\title{Deep Learning for Human Parsing: A Survey}
\author{Xiaomei Zhang, Xiangyu Zhu,~\IEEEmembership{Senior Member,~IEEE}, Ming Tang,~\IEEEmembership{Member,~IEEE}, and Zhen Lei,~\IEEEmembership{Senior Member,~IEEE}
\thanks{Xiaomei Zhang, Xiangyu Zhu and Ming Tang are with the National Laboratory of Pattern Recognition, Institute of Automation, Chinese Academy of Sciences, Beijing 100190,
China, and also with the School of Artificial Intelligence, University of Chinese Academy of Sciences, Beijing 100049, China (e-mail: xiaomei.zhang@nlpr.ia.ac.cn; xiangyu.zhu.chen@nlpr.ia.ac.cn; tangm@nlpr.ia.ac.cn).

Zhen Lei is with the National Laboratory of Pattern Recognition (NLPR),
Institute of Automation, Chinese Academy of Sciences (CASIA), Beijing
100190, China, also with the School of Artificial Intelligence, University of
Chinese Academy of Sciences (UCAS), Beijing 100190, China, and also with
the Centre for Artificial Intelligence and Robotics, Hong Kong Institute of
Science and Innovation, Chinese Academy of Sciences, Hong Kong, China
(e-mail: zlei@nlpr.ia.ac.cn).}

\thanks{Manuscript received Jan xx, 2023.} 
}
\markboth{IEEE TRANSACTIONS ON IMAGE PROCESSING, Jan. -, NO. -, - 2023}%
{Shell \MakeLowercase{\textit{et al.}}: Bare Demo of IEEEtran.cls for IEEE Journals}

\maketitle

\begin{abstract}
Human parsing is a key topic in image processing with many applications, such as surveillance analysis, human-robot interaction, person search, and clothing category classification, among many others. Recently, due to the success of deep learning in computer vision, there are a number of works aimed at developing human parsing algorithms using deep learning models. As methods have been proposed, a comprehensive survey of this topic is of great importance. In this survey, we provide an analysis of state-of-the-art human parsing methods, covering a broad spectrum of pioneering works for semantic human parsing. We introduce five insightful categories: (1) structure-driven architectures exploit the relationship of different human parts and the inherent hierarchical structure of a human body, (2) graph-based networks capture the global information to achieve an efficient and complete human body analysis, (3) context-aware networks explore useful contexts across all pixel to characterize a pixel of the corresponding class, (4) LSTM-based methods can combine short-distance and long-distance spatial dependencies to better exploit abundant local and global contexts, and (5) combined auxiliary information approaches use related tasks or supervision to improve network performance. We also discuss the advantages/disadvantages of the methods in each category and the relationships between methods in different categories, examine the most widely used datasets, report performances, and discuss promising future research directions in this area.
\end{abstract}


\IEEEpeerreviewmaketitle

\section{Introduction}

\begin{figure}[t]
 \begin{center}
 \includegraphics[width=8.9cm]{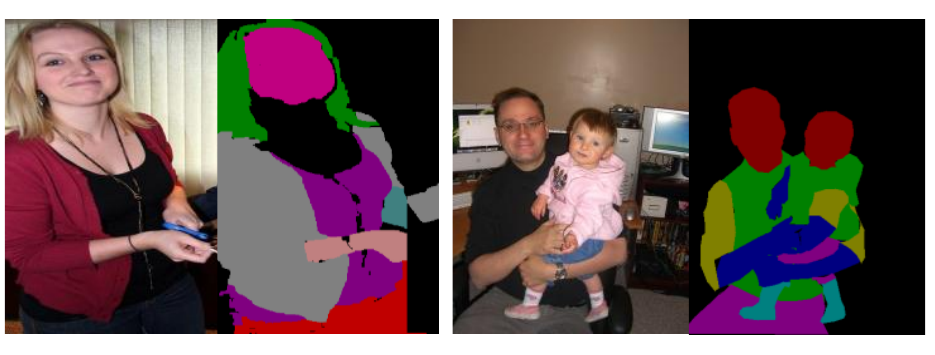}
 \caption{The goal of human parsing is to assign human images into multiple human parts of fine-grained semantics and benefit a detailed understanding of images. }
 \label{fig:sample}
 \end{center}
\end{figure}

\IEEEPARstart{H}{uman} parsing is an essential component in many visual understanding systems. It aims to assign human images into multiple human parts of fine-grained semantics and benefits a detailed understanding of images, some examples are shown in Fig.~\ref{fig:sample}. Human parsing plays a central role in a broad range of applications, including human-friendly robots~\cite{ulku2022survey}, person re-identification \cite{Farenzena2010Person}, human behavior analysis \cite{Wang2012Discriminative}, clothing style recognition and retrieval \cite{Yamaguchi2014Paper}, to name a few. Many human parsing algorithms have been developed in the literature, in the early stage, such as Markov random field~\cite{tompson2014joint}, support vector machines~\cite{bourdev2009poselets}, cascaded pictorial structures model~\cite{sapp2010cascaded}, part-based tree~\cite{wang2013beyond}, quadratic deformation cost~\cite{sapp2013modec}, binary random variable~\cite{kiefel2014human}. Over the past few years, deep learning for human parsing achieves remarkable performance improvements, especially, FCNet~\cite{long2015fully} was proposed.

In this paper, our motivation is to focus on the recent works in human parsing and discuss deep learning-based parsing 2D methods proposed until 2022. We provide a comprehensive review and insight on different aspects of these methods, including network architectures, training data, main contributions, and their limitations. Deep learning-based human parsing methods can be divided into the following categories based on their main technical contributions: (1) structure-driven architectures, (2) graph-based networks, (3) context-aware networks, (4) LSTM-based methods, (5) combined auxiliary information approaches contain pose-based auxiliary methods, edge-based auxiliary methods and detection-based auxiliary methods, and (6) other models. Additionally, we review some of the most popular human parsing datasets and their performance measures used for evaluating the methods. In the end, we discuss several potential future directions and applications for deep learning-based human parsing methods.

Some key contributions of our survey can be summarized as follows:
\begin{itemize}
  \item[1.] Our survey focuses on the recent papers with respect to human parsing, and overviews deep learning-based human parsing algorithms proposed until 2022.
  \item[2.] This survey provides insight to human parsing methods, including network architectures, training data, main contributions, and their limitations.
  \item[3.] We review popular human parsing datasets and provide a comparative summary of the properties and performance of the reviewed methods for parsing purposes, on popular benchmarks.
  \item[4.] We discuss several potential future directions and applications for deep learning-based human parsing methods.
\end{itemize}

\begin{table*}\Huge
\begin{center}
\caption{Taxonomy of Head Pose Estimation Approaches
} \label{t0}
\resizebox{1\textwidth} {!} {
\begin{threeparttable}
    \begin{tabular}{ c|c|c }
    \toprule
     Approach&Highlights&Representative Works  	\cr
    \midrule
    \midrule
    Structure-driven Architectures   &\makecell{Exploiting the relationship of different human parts
\\ and the inherent hierarchical structure of a human body.}  &A-AOG~\cite{park2018attribute}, PCNet~\emph{et al}.~\cite{zhu2018pcn}  \cr
    \midrule
    Graph-based Networks &\makecell{Capturing the global information to achieve
\\ an efficient and complete human body analysis.} &Graphonomy~\cite{ke2019graphonomy,lin2020graphonomy},CNIF~\cite{wang2019learning},HHP~\cite{wang2020hierarchical} \cr
    \midrule
    Context-aware Networks  &\makecell{Exploring useful contexts across all pixel
\\ to characterize a pixel of the corresponding class.} &ATR~\cite{Liang2015Deep}, M-CNN~\cite{liu2015matching}, Co-CNN~\cite{liang2015human}, SCHP~\cite{li2020self}   \cr
    \midrule
    LSTM-based Methods &\makecell{Combining short-distance and long-distance spatial dependencies
\\ to better exploit abundant local and global contexts.} &LG-LSTM~\cite{Liang2016Semantic}, Graph LSTM~\cite{liang2016semantic1}, structure-evolving LSTM~\cite{Liang2017Interpretable}   \cr
    \midrule
    \makecell{Combined Auxiliary \\ Information Approaches} &Using related tasks or
supervision to improve network performance. &  \makecell{Pose-based (JMPE~\cite{Xia2017Joint}, SSL~\cite{gong2017look}, MuLA~\cite{nie2018mutual}.), \\ Edge-based ( CorrPM~\cite{9157207}, CE2P~\cite{Liu2019CE2P}), \\ Detection-based (HAZN~\cite{Xia2015Zoom}, Li~\emph{et al}.~\cite{li2017holistic})}  \cr
    \bottomrule
    \end{tabular}
\end{threeparttable}
}
\end{center}
\end{table*}

The remainder of the paper is organized as follows: Section 2 provides a comprehensive overview of deep learning-based human parsing methods according to their main technical contributions. We also discuss their strengths and limitation. Section 3 introduces some of the most popular human parsing datasets and evaluation metrics. Section 4 discusses several applications and potential future directions. Finally, conclusions are given in Section 5.


\section{Deep Learning-Based Human Parsing}
This section provides a detailed review of deep learning-based human parsing methods proposed until 2022. We grouped these methods into five categories based on their model architecture. Table \ref{t0} provides a list of representative systems for each of these categories.

\begin{figure}[h]
 \begin{center}
 \includegraphics[width=7.8cm]{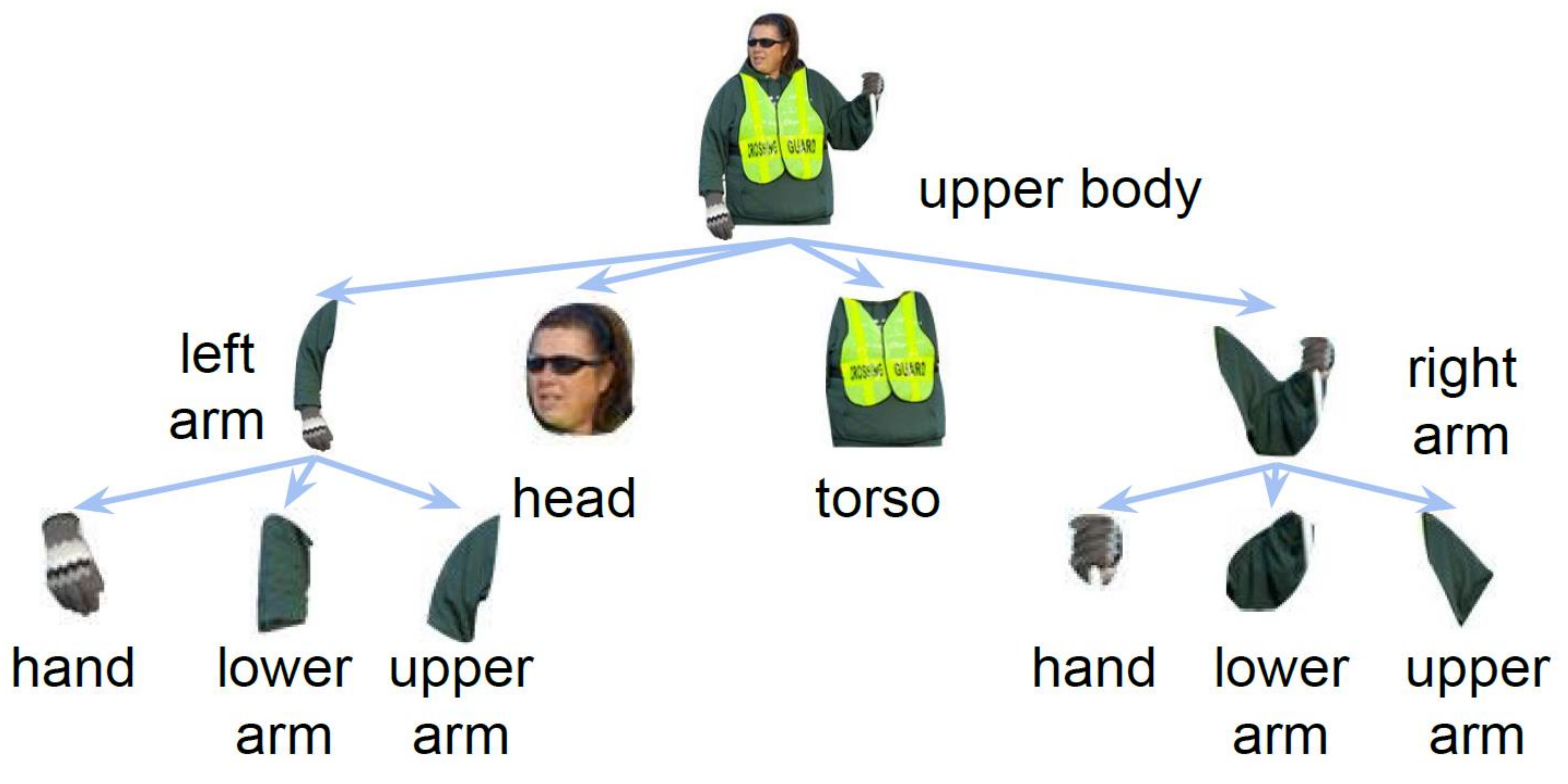}
 \caption{Phrase structure grammar is based on the constituency grammar which defines the rule to break down a node into its constituent parts. From~\cite{park2018attribute}. }
 \label{fig:structure grammar}
 \end{center}
\end{figure}

\subsection{Structure-Driven Architectures}
The human body is a natural hierarchy, how to use the prior knowledge to segment different human parts is an urgent problem to be solved. Park~\emph{et al}.~\cite{park2018attribute} present an attribute and-or grammar (A-AOG) model for inferring human parts in the hierarchical representation, which also jointly represents the human pose and human attribute. A-AOG explicitly represents the decomposition and articulation of body parts and accounts for the correlations between parts. The network uses phrase structure grammars to represent the hierarchical decomposition of the human body from whole to parts, and employs dependency grammars to model the geometric articulation.

\begin{figure}[h]
 \begin{center}
 \includegraphics[width=8.8cm]{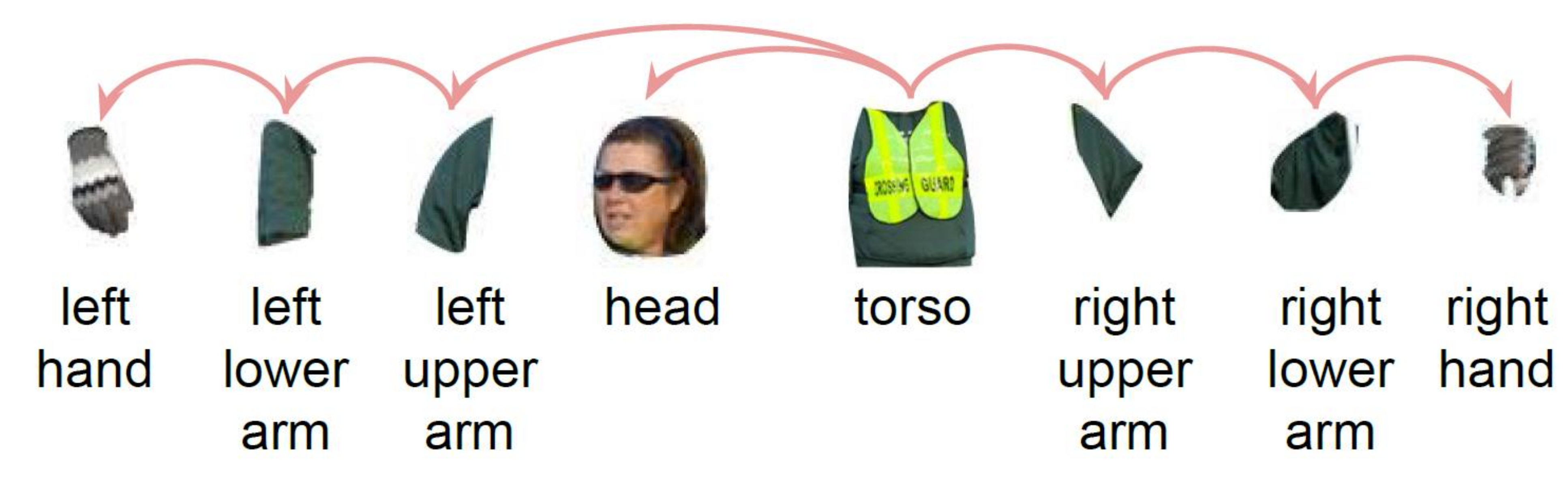}
 \caption{Dependency grammar defines adjacency relations that connect the geometry of a part to its dependent parts. From~\cite{park2018attribute}.}
 \label{fig:dependency grammar}
 \end{center}
\end{figure}

The phrase structure grammar represents human parts in a coarse-to-fine method based on the constituency relation. The grammar is defined as:
\begin{equation}
\label{eq:mm}
\begin{split}
&a\rightarrow a_1|a_2|a_3,\\
\end{split}
\end{equation}
where a typical nonterminal node $a\in v_n$, $a_i$ is a string of nodes in $v_n\cup v_t$, $v_n, v_t$ denotes human parts. Fig.~\ref{fig:structure grammar} shows an example. The root node is the upper body and decomposed into arms, head, and torso. The arms are further decomposed into upper-arm, lower-arm, and hand.
Dependency grammars have been widely used in natural language processing for syntactic parsing. Fig.~\ref{fig:dependency grammar} is a parsing grammar for the upper body. The root node is the torso part as it is the center of the body and connected to other parts. The upper arms and head are the child nodes of the torso.

\begin{figure}[h]
 \begin{center}
 \includegraphics[width=8.8cm]{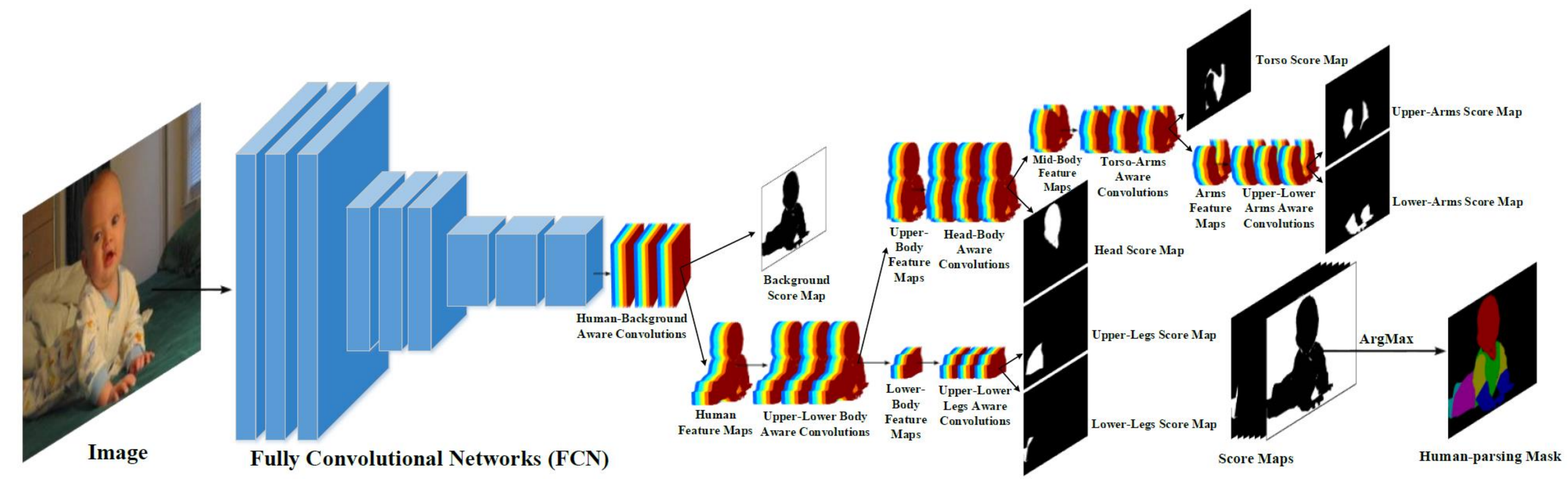}
 \caption{The framework of PCNet. From~\cite{zhu2018pcn}.}
 \label{fig:pcnet}
 \end{center}
\end{figure}

Zhu~\emph{et al}.~\cite{zhu2018pcn} propose a progressive cognitive structure to segment human parts. In the hierarchical network, the latter layers inherit information from former layers and pay attention to a finer component. As shown in Fig.~\ref{fig:pcnet}, the given image is sent into a FCN to extract original features. And then the image-level (original) features are decomposed into a background score map and human-level features. The human-level features are further decomposed into upper-body features and lower-body features, repeating the above steps until all are segmented.

The structure-driven human parsing methods explore the inherent relations of human parts according to the prior knowledge of the human body. These methods can make the network pay more attention to the interested human body itself and reduce the interference of background and redundant information. However, these methods are easy to lead to error accumulation. For example, if the root node is wrong, the error will be passed to the subsequent nodes, resulting in error accumulation.

\subsection{Graph-Based Networks}
\begin{figure}[h]
 \begin{center}
 \includegraphics[width=8.8cm]{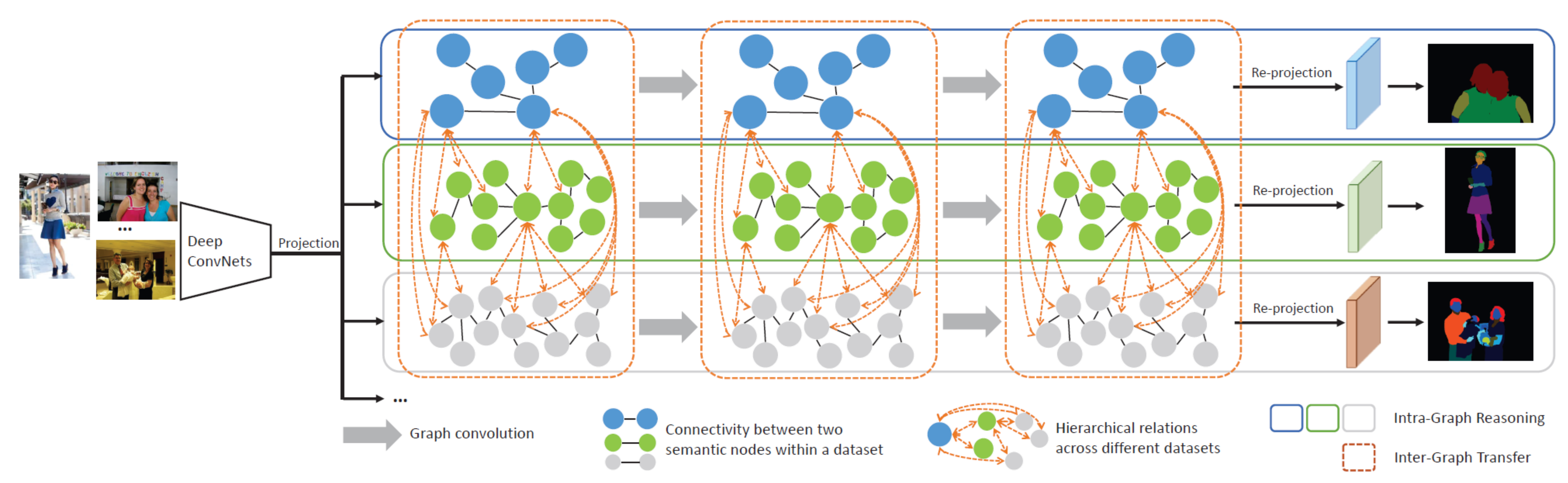}
 \caption{The framework of Graphonomy. From~\cite{ke2019graphonomy}.}
 \label{fig:graphonomy}
 \end{center}
\end{figure}

The graph convolution~\cite{bruna2013spectral} can effectively explore the semantic relationship among human parts, thus, some works~\cite{ke2019graphonomy,lin2020graphonomy} introduce the graph convolution into the human parsing task.
As shown in Fig.~\ref{fig:graphonomy}, Gong~\emph{et al}.~\cite{ke2019graphonomy,lin2020graphonomy} design a graph-based human parsing network, named "Graphonomy", which employs graph convolution to capture the global information and semantic consistency. The image features extracted by the deep convolution network are projected into a high-level graph representation, where the body parts are nodes and the relationships between the parts are edges.
Graphonomy first learns and propagates compact high-level graph representation among parts within one dataset via intra-graph reasoning, and then transfers semantic information across different datasets via inter-graph transfer. In this way, Graphonomy takes advantage of different granular annotated data.
However, multiple datasets are required in the training process, which consumes many computing resources.

Wang~\emph{et al}.~\cite{wang2019learning} propose a method to explore the structural hierarchy of the human body by using a graph convolutional network. This method achieves an efficient and complete human body analysis, named an information fusion framework. This model models three inference processes: direct inference (directly predicting each part of a human body using image information), bottom-up inference (assembling knowledge from constituent parts), and top-down inference (leveraging context from parent nodes).

\begin{figure}[h]
 \begin{center}
 \includegraphics[width=8.8cm]{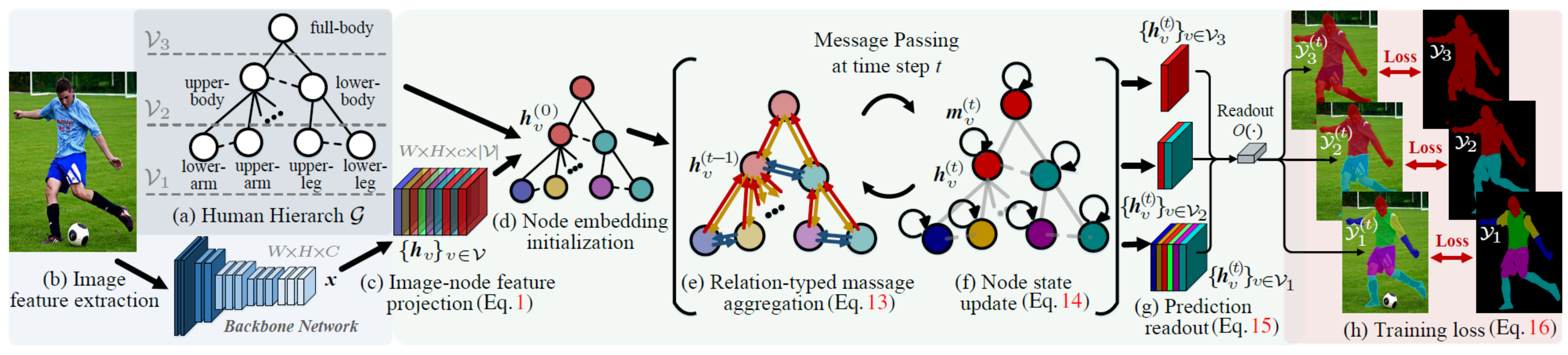}
 \caption{Illustration of the hierarchical graph for the human parsing task. From~\cite{wang2020hierarchical}.}
 \label{fig:hierarchical_graph}
 \end{center}
\end{figure}

In addition, Wang~\emph{et al}.~\cite{wang2020hierarchical} represent a hierarchical graph for the human parsing task. There are three kinds of part relations, i.e., decomposition, composition, and dependency, which are completely and precisely described by three distinct graph networks. This method represents the human semantic structure as a directed, hierarchical graph $g=(\upsilon, \varepsilon,y)$. As shown in Fig.~\ref{fig:hierarchical_graph}, the node set $\upsilon=\bigcup^3_{l=3}\upsilon_l$ denotes human parts in three different semantic levels, including the leaf nodes $\upsilon_1$, two middle-level nodes $\upsilon_2$ and one root $\upsilon_3$. The edge set $\varepsilon\in(^\upsilon_2)$ represents the relations between human parts (nodes), i.e., the directed edge. $y$ represents the ground truth maps. Given an input image, a fully convolution network first extracts image features projected into node (part) features. Then node features are sent into a hierarchical graph to capture expressive relation information and predicted different granularity of results.
The whole network is trained in graph learning methods and the supervision is human parsing datasets.

In graph-based human parsing networks, the semantic relationships of different parts are introduced into the parsing models, which can amend the accumulated errors by communicating with each other. In general, the body parts are the nodes and the relationship between the parts is the edge. Due to the diversity of the scale, occlusion, deformation and posture of human body parts, this kind of method is difficult to obtain rich contextual information, which affects the recognition ability of the model.

\begin{figure}[h]
 \begin{center}
 \includegraphics[width=8.8cm]{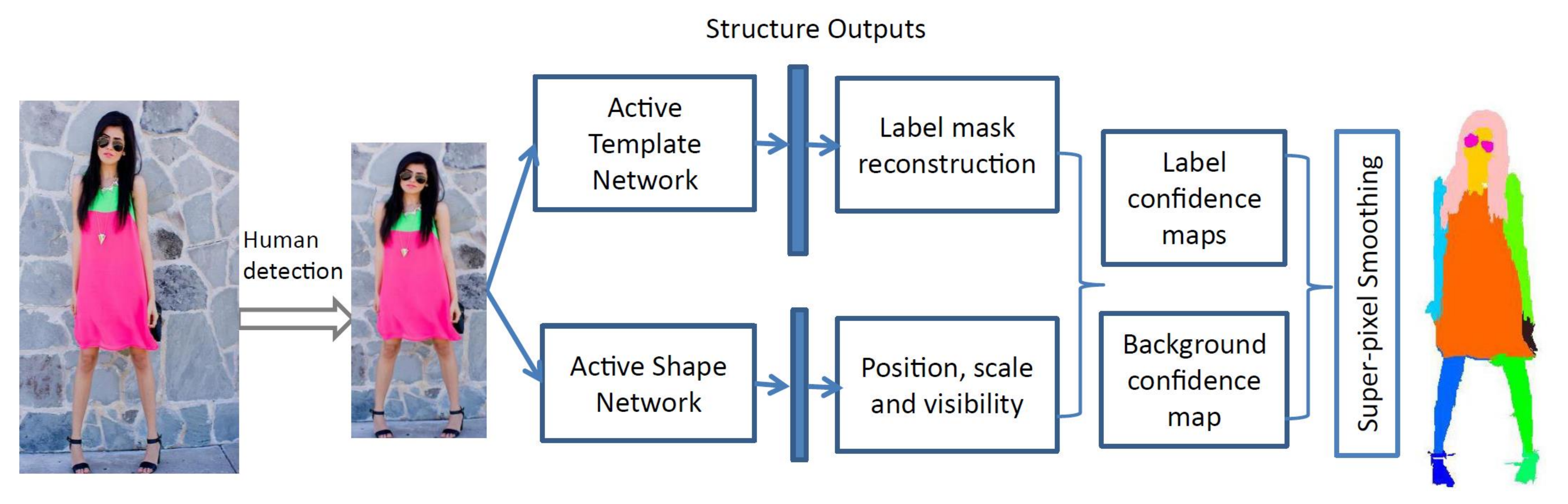}
 \caption{The framework of the active template regression model. From~\cite{Liang2015Deep}.}
 \label{fig:active_template}
 \end{center}
\end{figure}

\subsection{Context-Aware Networks}
Benefiting from the development of advanced CNN architectures and training techniques as well as the recently released human parsing datasets, context-aware methods attract more and more attention. As shown in Fig.~\ref{fig:active_template}, Liang~\emph{et al}.~\cite{Liang2015Deep} are the first to apply a convolutional neural network to the human parsing task, called active template regression model. The method is built on the end-to-end relation between the input human image and the structure outputs for human parsing. The first CNN network (active template network) is with max-pooling and designed to predict the template coefficients for each label mask, and the second (active shape network) is without max-pooling to preserve sensitivity to label mask position and generate the active shape parameters. Given an image, the outputs of the two networks are fused to generate the probability of each category for each pixel, and superpixel smoothing is finally used to refine the human parsing prediction. Extensive experiments demonstrate the significant superiority of the ATR framework over other contemporaneous states of the arts for human parsing.

\begin{figure}[t]
 \begin{center}
 \includegraphics[width=6.8cm]{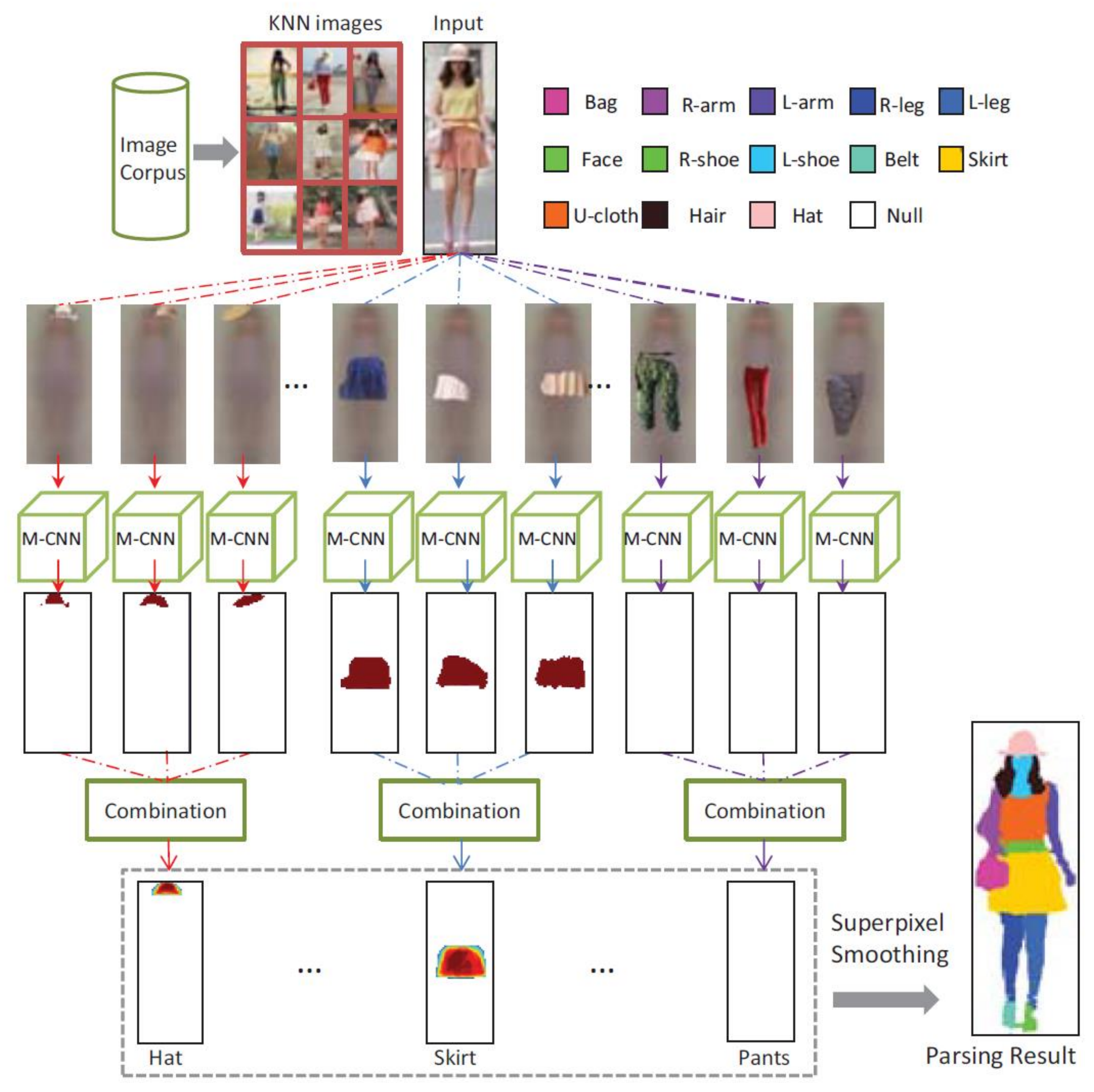}
 \caption{The architecture of the quasi-parametric network. From~\cite{liu2015matching}.}
 \label{fig:matchingcnn}
 \end{center}
\end{figure}

Subsequently, Liang~\emph{et al}.~\cite{liu2015matching} design a quasi-parametric method, which takes advantage of both parametric and non-parametric approaches, namely supervision from annotated data and the flexibility to use newly annotated images. Under the classic K Nearest Neighbor (KNN)-based nonparametric framework, the parametric Matching Convolutional Neural Network (M-CNN) is proposed to predict segment results. As shown in Fig.~\ref{fig:matchingcnn}, given a testing image, its KNN images are retrieved from the manually-annotated image corpus. Then the input image is paired with each semantic region of its KNN images and each pair is fed into the M-CNN individually. The M-CNN predicts the matching confidence and displacements between the input image pair. All corresponding label maps are combined to generate a probability map for each pixel which is further refined by superpixel smoothing.

\begin{figure}[h]
 \begin{center}
 \includegraphics[width=8.8cm]{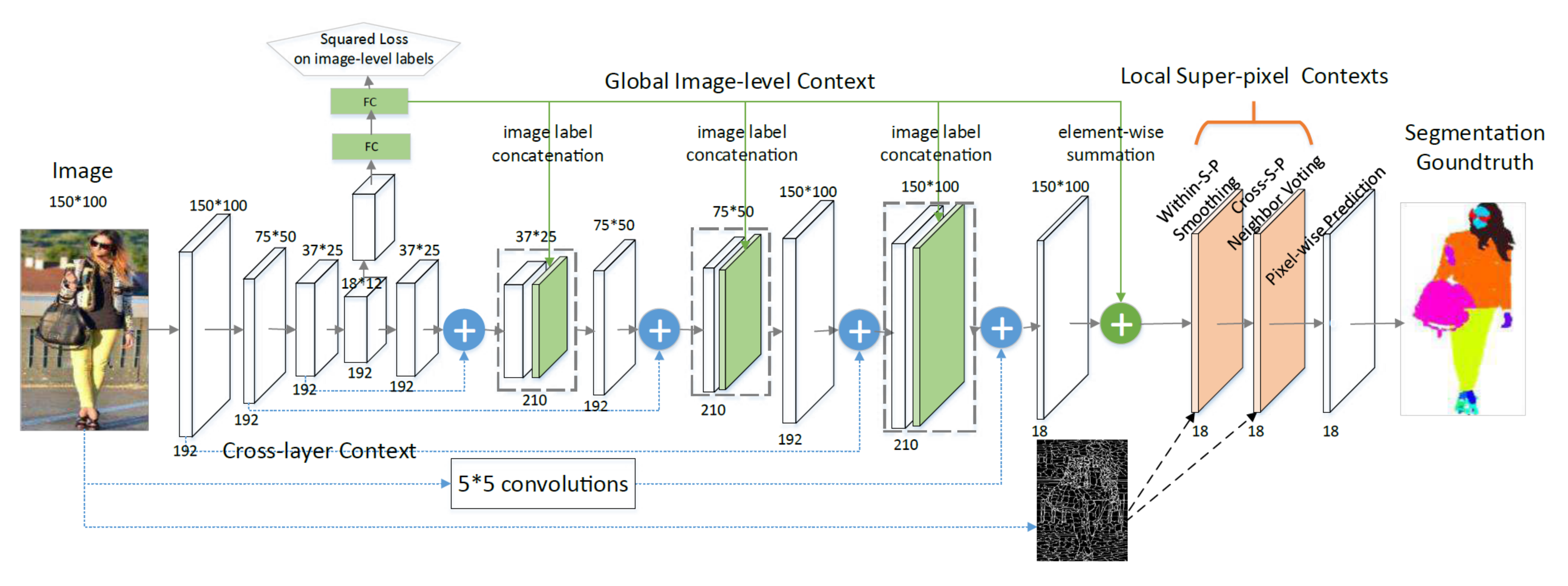}
 \caption{The framework of Contextualized Convolutional Neural Network. From~\cite{liang2015human}.}
 \label{fig:Co-CNN}
 \end{center}
\end{figure}

To effectively integrate all kinds of contexts into a unified model in the human parsing task, Liang~\emph{et al}.~\cite{liang2015human} propose a novel Contextualized Convolutional Neural Network (Co-CNN). This network integrates the cross-layer context, global image-level context, within-super-pixel context and cross-super-pixel neighborhood context to obtain rich context, improving the performance of human parsing (Fig.~\ref{fig:Co-CNN}). The cross-layer context is captured by the basic structure, which hierarchically combines the global semantic and local details across different layers. Then, the global image-level context serves as an auxiliary object in the intermediate layer and guides the subsequent feature learning. Finally, the local super-pixel contexts, the within-super-pixel context, and the cross-super-pixel neighborhood context are formulated as natural subcomponents to achieve local label consistency.

\begin{figure}[t]
 \begin{center}
 \includegraphics[width=8.8cm]{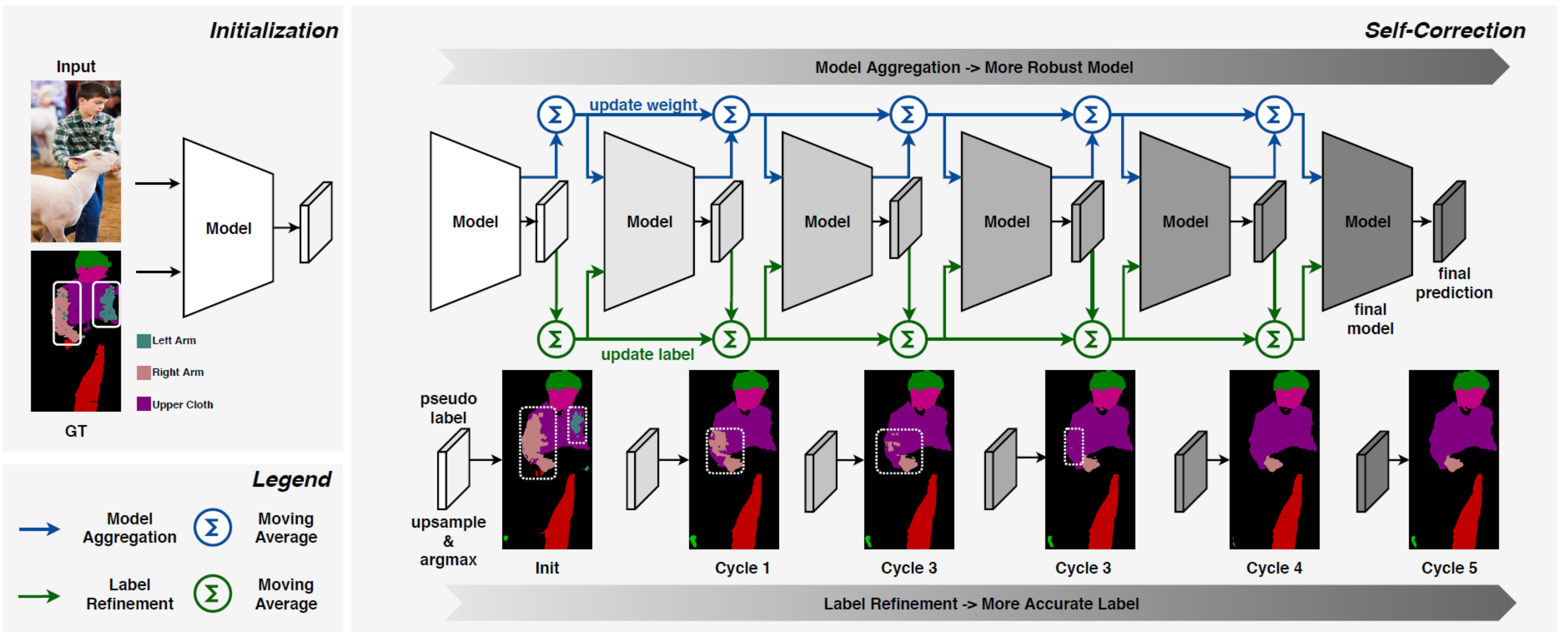}
 \caption{Framework of the Self-Correction for Human Parsing. From~\cite{li2020self}.}
 \label{fig:schp}
 \end{center}
\end{figure}

Li~\emph{et al}. design a noise-tolerant method, named Self-Correction
for Human Parsing (SCHP)~\cite{li2020self}, to progressively promote the reliability of the supervised labels as well as the learned models. As shown in Fig.~\ref{fig:schp}, SCHP first takes a model trained with inaccurate labels as initialization and then uses a cyclical learning scheduler to infer more reliable pseudo masks by iteratively aggregating the models in an online manner. Besides, the corrected labels in turn boost the model's performance.
In this way, the models and the labels will reciprocally become more robust and accurate.
The human parsing methods based on contexts can obtain rich semantics and details, expanding the effective receptive field. However,  this approach may result in information decay because the network is too deep.

\subsection{LSTM-based Methods}
\begin{figure}[h]
 \begin{center}
 \includegraphics[width=8.8cm]{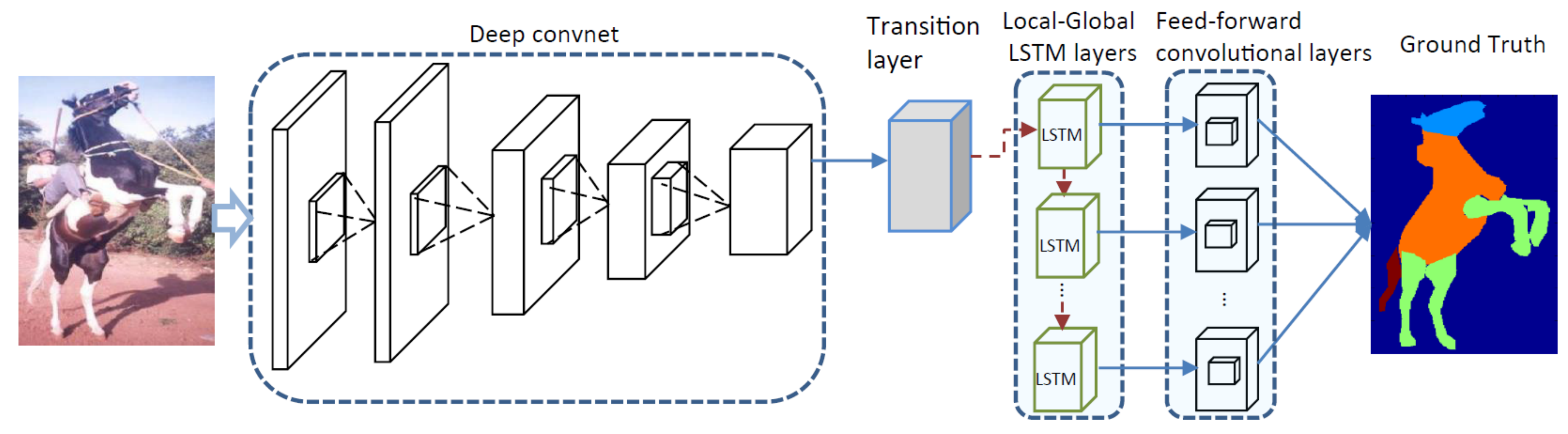}
 \caption{Architecture of the LG-LSTM network. From~\cite{Liang2016Semantic}.}
 \label{fig:lg-LSTM}
 \end{center}
\end{figure}
Long short-term memory (LSTM)~\cite{he2016reading} uses memory cells to independently read, write and forget some information, retaining the main information. It can mitigate the decay of information.
Thus, Liang~\emph{et al}. introduce LSTM into the human parsing task. First, Liang~\emph{et al}.~\cite{Liang2016Semantic} propose a novel Local-Global Long Short-Term Memory (LG-LSTM) to combine short- and long-distance spatial dependencies (Fig.~\ref{fig:lg-LSTM}). LG-LSTM layer obtains local guidance from neighboring positions and global guidance from the whole image and then imposes on each position to better exploit abundant local and global contexts. Given an input image, the backbone obtains its original features. Then, these features are sent into the transition layer and several stacked LG-LSTM layers to improve the ability to feature. The feed-forward convolutional layers are appended to generate the prediction.

Then, Liang~\emph{et al}. further design Graph Long Short-Term Memory (Graph LSTM)~\cite{liang2016semantic1} to improve LSTM, which is the generalization of LSTM from sequential data or multi-dimensional data to general graph-structured data. Graph LSTM takes each arbitrary-shaped super-pixel as a semantically consistent node, and adaptively constructs an undirected graph for each image, where the spatial relations of the super-pixels are naturally used as edges. The node updating sequence for Graph LSTM layers is determined by the confidence-drive scheme, and then Graph LSTM layers can sequentially update the hidden states of all super-pixel nodes.

\begin{figure}[h]
 \begin{center}
 \includegraphics[width=8.8cm]{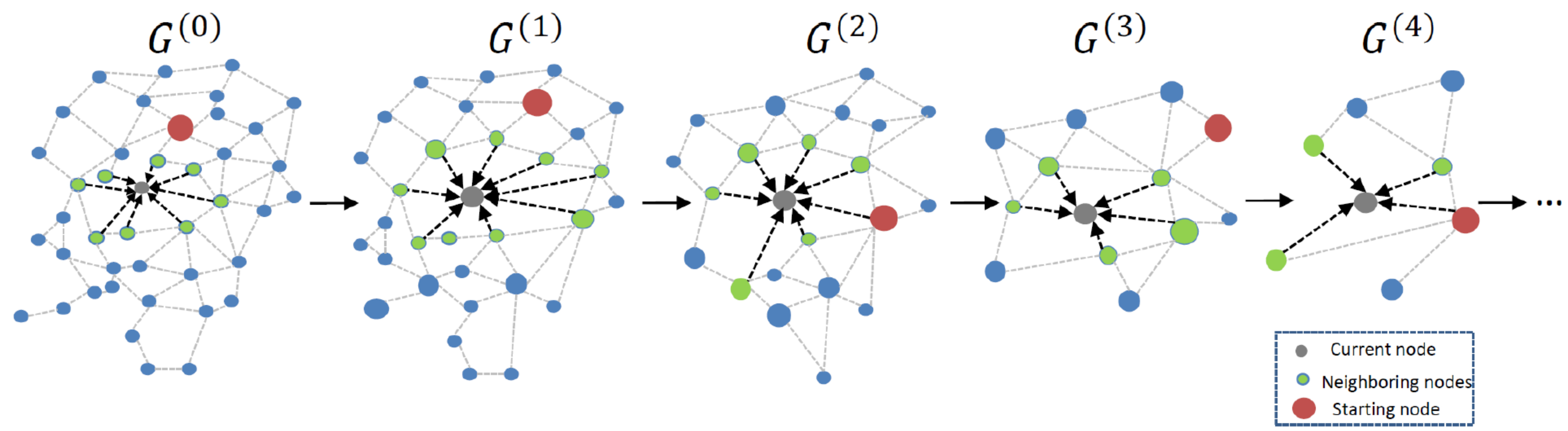}
 \caption{An illustration of the structure-evolving LSTM. From~\cite{Liang2017Interpretable}.}
 \label{fig:structure-evolrlstm}
 \end{center}
\end{figure}

In another work, Liang~\emph{et al}. propose to learn the intermediate interpretable multi-level graph structures in a progressive way, named structure-evolving LSTM~\cite{Liang2017Interpretable} (Fig.~\ref{fig:structure-evolrlstm}). In each LSTM layer, we estimate the compatibility of two connected nodes from their corresponding LSTM gate outputs, which is used to generate a merging probability. Then, the candidate graph structures are accordingly generated where the nodes are grouped into cliques with their merging probabilities.
The network produces the new graph structure with a Metropolis-Hasting algorithm, which alleviates the risk of getting stuck in local optimums by stochastic sampling with an acceptance probability. Once a graph structure is accepted, a higher-level graph is then constructed by taking the partitioned cliques as its nodes.

\begin{figure}[h]
 \begin{center}
 \includegraphics[width=8.8cm]{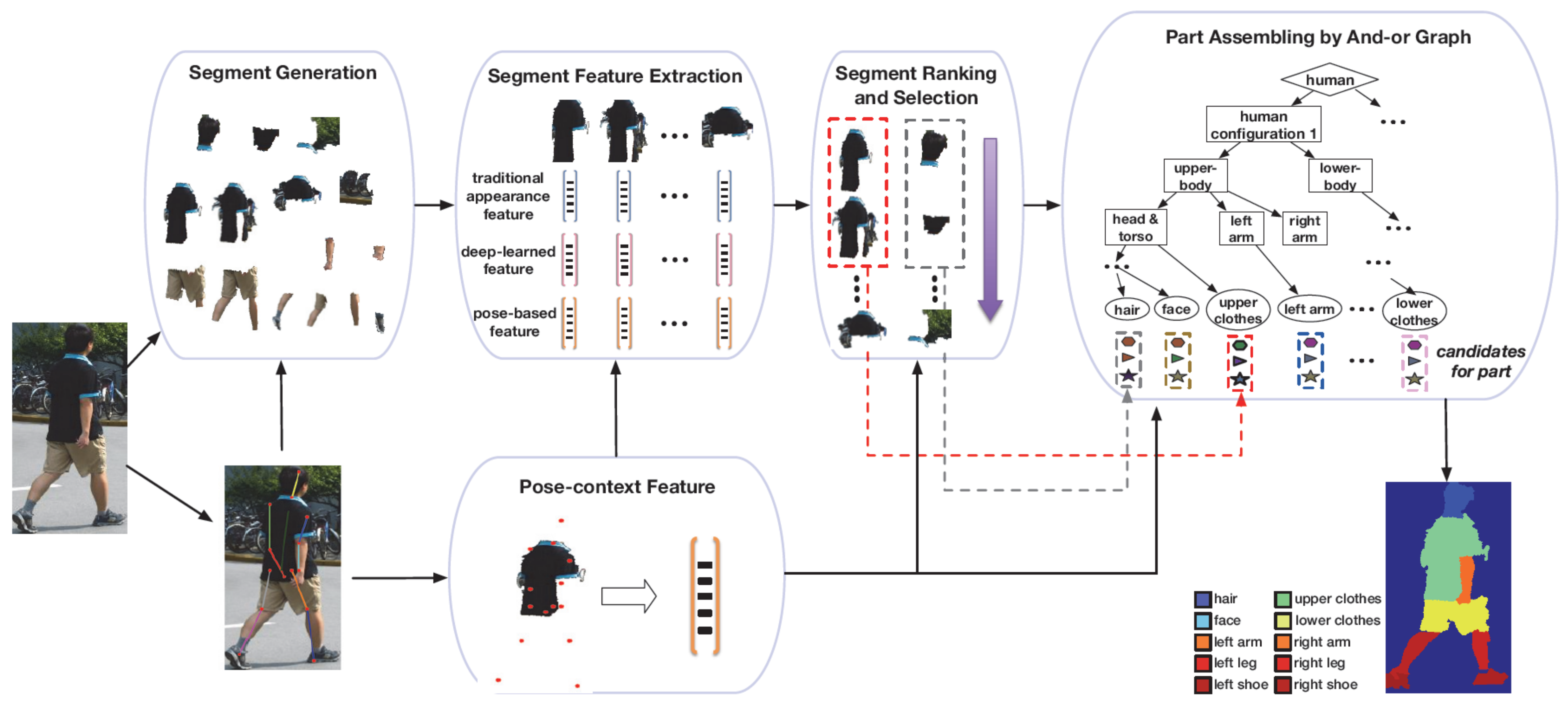}
 \caption{Pose-guided human parsing by an and-or graph using pose-context features for human parsing. From~\cite{xia2016pose}.}
 \label{fig:pose-guide-and-or-graph}
 \end{center}
\end{figure}

\subsection{Combined Auxiliary Information Approaches}
Human parsing is a pixel-level fine classification task, which requires rich and effective feature expression. Combining with human body detection, human posture estimation and human body edge information can greatly improve the expression ability of features and the performance of human body analysis.

\textbf{Pose-based Auxiliary Methods.}
In the computer vision community, human parsing and human pose estimation are two complementary tasks. Joints can provide object-level shape information for human parsing, and pixel-level analysis can constrain the changes of pose position.

Xia~\emph{et al}. propose a human parsing approach which uses human pose location as cues to provide pose-guided segment proposals for semantic parts~\cite{xia2016pose}, as shown in Fig.~\ref{fig:pose-guide-and-or-graph}. These segment proposals are ranked using standard appearance cues, called pose-context. These proposals are selected and assembled using an and-or-graph to output a parse of the person, which can deal with large human appearance variability. Given a pedestrian image, the network first uses a pose estimation to obtain human pose joints. Then, based on the joints and modified RIGOR algorithm~\cite{humayun2014rigor}, segments aligned with object boundaries are generated. Human pose joints further obtain pose-context feature which combines segments to extract the segment features. The segment proposal model selects and ranks the parts. Finally, part assembling generates the paring results. Based on this method~\cite{xia2016pose}, Xia~\emph{et al}. further improve human parsing methods based on human pose estimation and design a network~\cite{Xia2017Joint}. This network trains two fully convolutional neural networks, to optimize both tasks simultaneously.

\begin{figure}[h]
 \begin{center}
 \includegraphics[width=8.8cm]{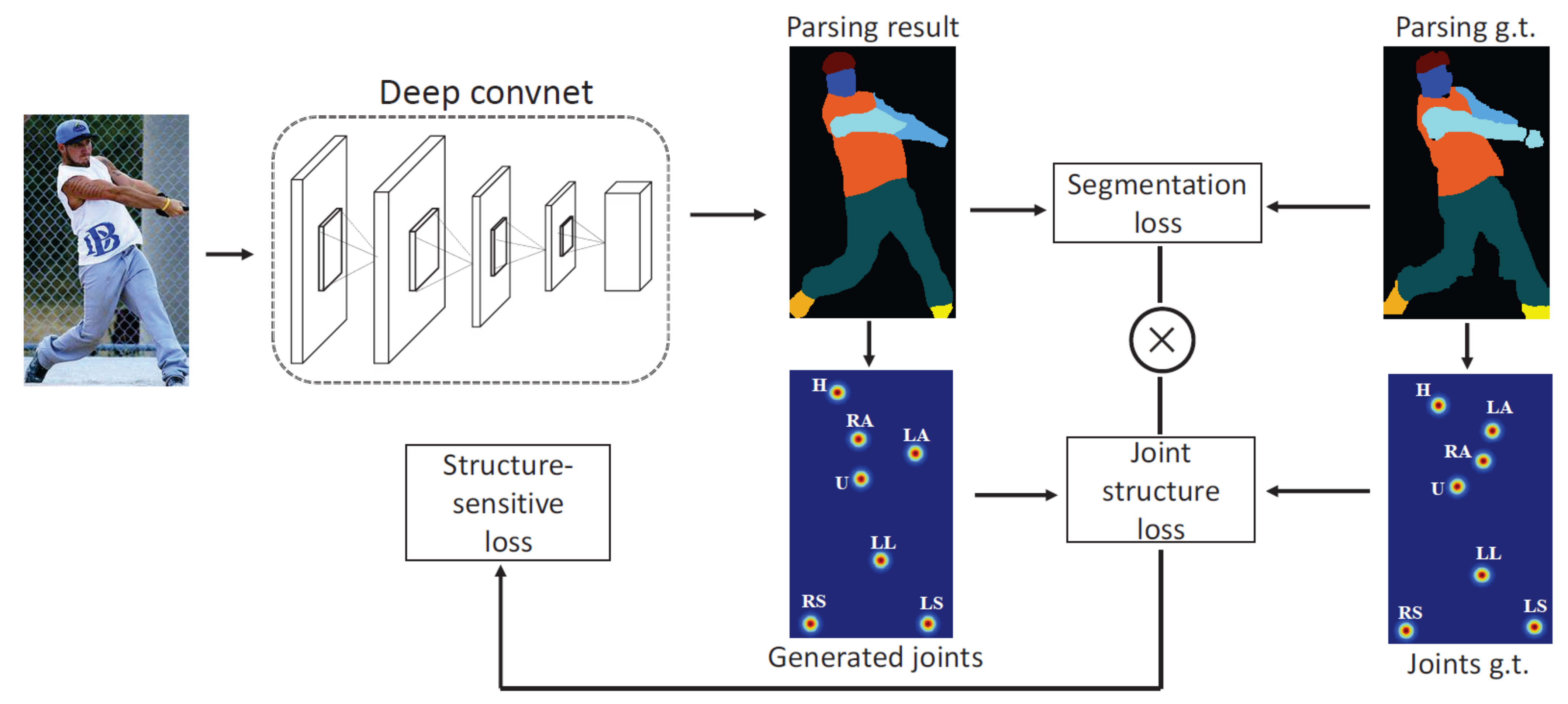}
 \caption{Framework of Self-supervised Structure-sensitive Learning (SSL) for human parsing. From~\cite{gong2017look}.}
 \label{fig:lip}
 \end{center}
\end{figure}

Gong~\emph{et al}. develop a novel Self-supervised Structure-sensitive Learning (SSL) approach~\cite{gong2017look}, which imposes human pose structure into parsing without resorting to extra supervision (Fig.~\ref{fig:lip}). Note that there is no need to specifically label human joints in model training. This framework can be injected into any advanced networks to help incorporate rich high-level knowledge from joint and improve the parsing results. Given an image, the backbone generates parsing results. The generated joints and joints labels are obtained by computing the center points of corresponding regions in parsing maps. The structure-sensitives loss is generated by weighting segmentation loss with joint structure loss.

Subsequently, Nie~\emph{et al}. present a Mutual Learning to Adapt model (MuLA) for joint human parsing and pose estimation~\cite{nie2018mutual}. MuLA predicts dynamic task-specific model parameters via recurrently leveraging guidance information from its parallel tasks. Thus MuLA combines the advantages of parsing and pose models to provide more powerful representations by incorporating information from their counterparts, generating more accurate results.

These two tasks can improve each other's feature representation and enhance the accuracy of models. However, the two tasks have different concerns. Human pose estimation mainly focuses on global features and cannot provide a unique label for each pixel. Therefore, the help to the human parsing task is limited. At the same time, the two optimization methods are different, which will affect the improvement of accuracy.

\textbf{Edge-based Auxiliary Methods.}
In the image, the low-frequency region is the area with similar semantics, while the high-frequency region is usually the area with a larger semantic transformation ratio of area block. The edge of the human body is used to describe the area of human foreground and background transformation, which has regional differentiation. On the basis of Co-CNN~\cite{liang2015human}, Liang~\emph{et al}~\cite{Liang2016Human}. add a branch to guide the separation of the foreground and the background regions by information on the edge of the human body, which improves the recognition ability of the model to different semantic regions. An advanced semantic boundary is used to guide pixel-level labeling and improve the accuracy of the model.

\begin{figure}[t]
 \begin{center}
 \includegraphics[width=7.0cm]{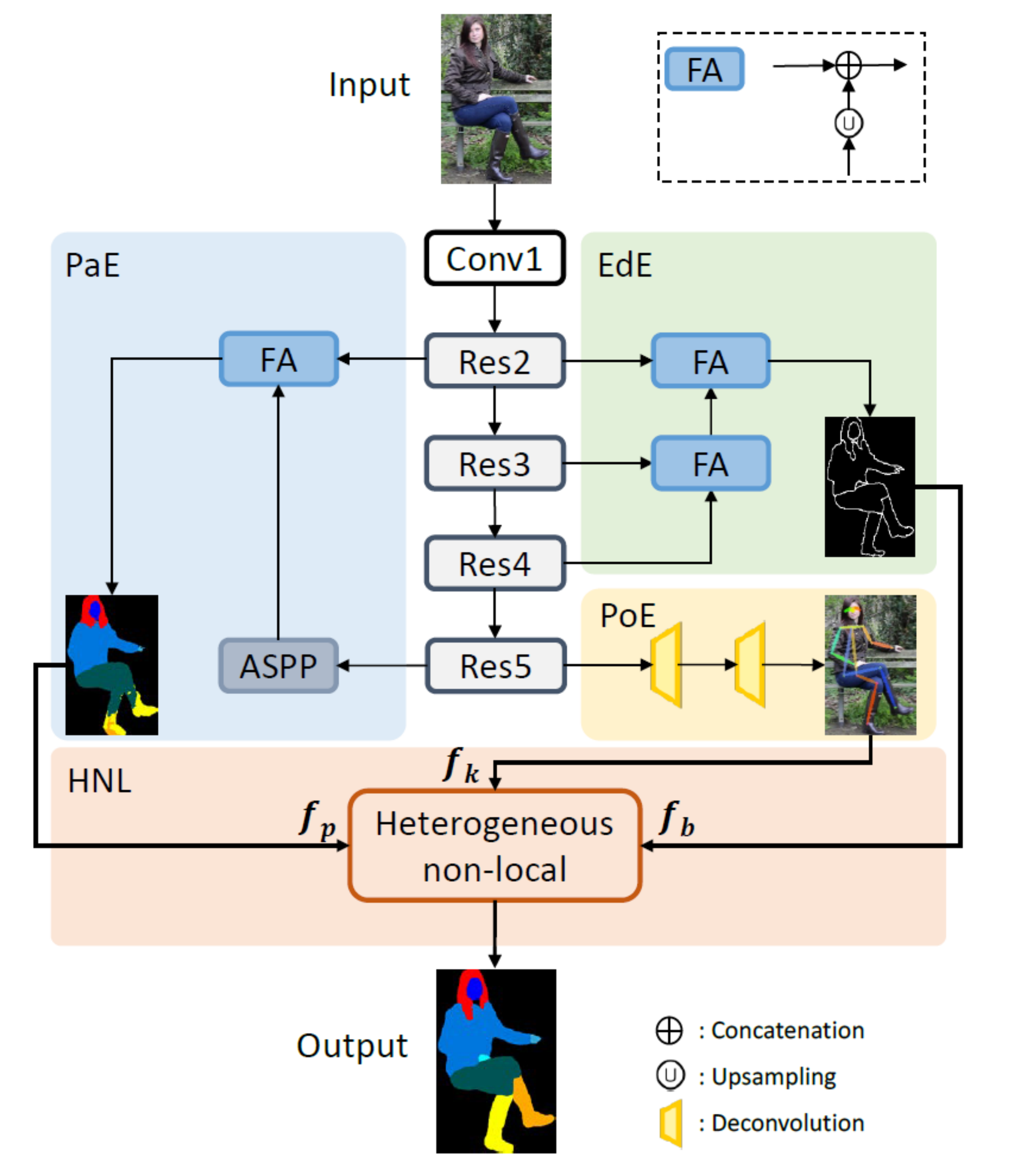}
 \caption{Framework of the Correlation Parsing Machine (CorrPM) framework. From~\cite{9157207}.}
 \label{fig:CorrPM}
 \end{center}
\end{figure}

Zhang~\emph{et al}. propose a Correlation Parsing Machine (CorrPM)~\cite{9157207} to study how human semantic boundaries and keypoint locations can jointly improve human parsing (Fig.~\ref{fig:CorrPM}). CorrPM uses a heterogeneous non-local block to discover the spatial affinity among feature maps from the edge, pose and parsing. The input images go through the backbone to generate features in different stages. Features of the second and fifth stages are sent into the paring encoder to obtain coarse segmentation maps. Features of the second, third and fourth stages are sent into the edge encoder to generate semantic boundaries. Features of the five stages are sent into the pose encoder to predict joint location. The heterogeneous non-local is appended to explore the correlation among the three factors and generates fine results.

Liu~\emph{et al}.~\cite{Liu2019CE2P} design a simple yet effective context embedding with edge perceiving (CE2P) framework, which combines feature resolution, global context, and edge details.
CE2P brings a high-performance boost to single human parsing task and serve as a solid baseline for future research in single/multiple human parsing.

The edge label of the human body can be obtained through the edge processing (e.g., Canny~\cite{canny1986computational}) of the parsing label. This kind of natural information does not need additional labeling and is cost-effective. But the number of pixels occupied by the edges is small compared to the overall image. At present, the average intersection ratio is usually used to evaluate the performance of the model, and the edge accuracy has little advantage in this calculation method. Usually, the effect of the model edge has been greatly improved, but the analytical accuracy of human parsing has not been improved. Therefore, edge information cannot effectively improve the evaluation value of network performance.

\begin{figure}[h]
 \begin{center}
 \includegraphics[width=8.0cm]{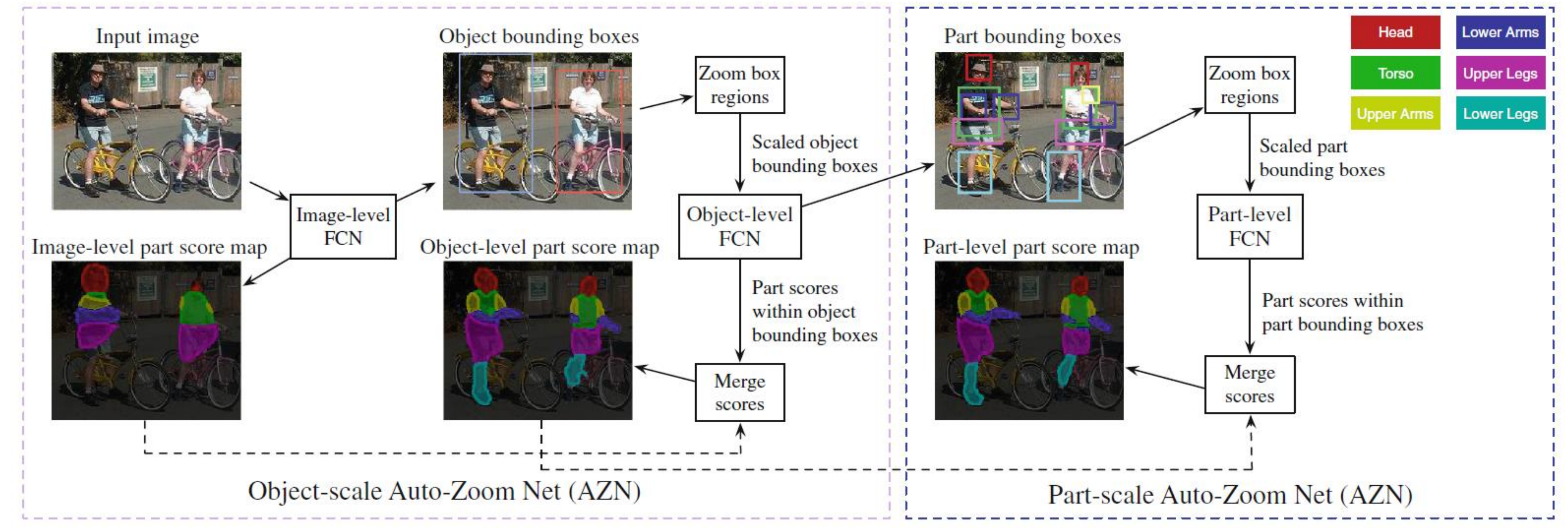}
 \caption{Framework of the Hierarchical Auto-Zoom Net (HAZN) framework. From~\cite{Xia2015Zoom}.}
 \label{fig:hazn}
 \end{center}
\end{figure}
\textbf{Detection-based Auxiliary Methods.}
Due to the interference of complex scenes and similar backgrounds, it is difficult to extract complete and accurate foreground. Detecting the human body can reduce the interference of background information, making the network pay more attention to the foreground, and improving the precision of network analysis. As shown in Fig.~\ref{fig:hazn}, Xia~\emph{et al}.~\cite{Xia2015Zoom} design a Hierarchical Auto-Zoom Net (HAZN) for object parsing which adapts to the local scales of objects and parts. HAZN is a sequence of two Auto-Zoom Nets (AZNs), each one has two tasks: the first is to predict the locations and scales of object instances (the first AZN) or their parts (the second AZN); the second is to estimate the part scores for predicted object instance or part regions. Given an image, its part scores are predicted and refined by three FCNs with three levels of granularity, i.e., image-, object-, and part-level. At each level, the FCN outputs the score maps and generates the locations and scales for the next level.

\begin{figure*}[t]
 \centering
 \includegraphics[width=0.99\linewidth]{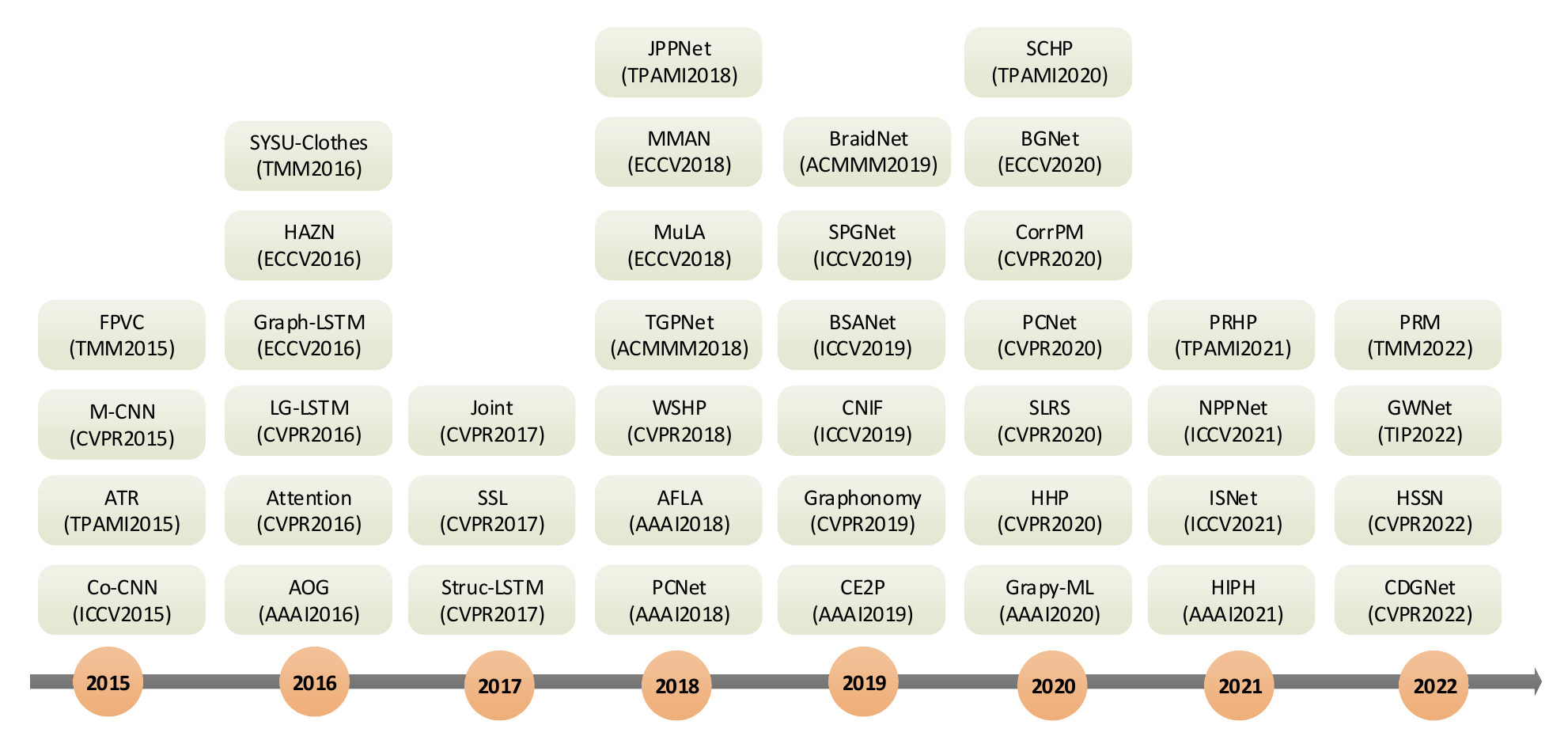}
 \caption{The timeline of deep learning-based human parsing algorithms, from 2015 to 2022.}
 \label{fig:timeline}
\end{figure*}

Li~\emph{et al}.~\cite{li2017holistic} address the human parsing task by segmenting the parts of objects at an instance level, such that each pixel in the image is assigned a part label, as well as the identity of the object it belongs to. An input image is sent into a human detection  network and a body parts semantic segmentation network, producing $D$ detections of human and parsing results, respectively. These results are used to form the unary potentials of an Instance CRF which performs instance segmentation by associating labeled pixels with human detections.

The detection methods can reduce the interference of background information and make the network more focused on the human body, but this method also has two drawbacks. On the one hand, even if the detection method is completely correct, the interference of background information cannot be completely avoided, and human parsing methods are still needed to extract the foreground from the background. When analyzing the results, the closer the foreground was to the body itself, the more disturbing it was. On the other hand, if the detection method is incorrect, subsequent human parsing will inherit the error, resulting in error accumulation. Therefore, the human body analysis method based on the detection model has some drawbacks.

\subsection{Other Models}
In addition to the above methods, there are several other deep learning methods for human parsing, such as the following: CDGNet~\cite{liu2022cdgnet} simplifies the complex spatial human parsing problem into the horizontal and vertical positions of human parts individually. Accordingly, the method builds the class distribution labels in the horizontal and vertical directions as new supervision signals from the original label of human parsing. Liu~\emph{et al}.~\cite{liu2020hybrid} propose a Hybrid Resolution Network (HyRN) by adding deconvolution and multi-scale supervision on top of the HRNet with minimal extra computation overhead. HyRN aggregates rich high-resolution representations to predict more accurate parsing results, especially for small components, on which the improvement to HRNet baseline exceeds 4 pp.
Lin~\emph{et al}.~\cite{lin2019cross} introduce an effective framework, called-domain complementary learning with a pose, to leverage information in both real and synthetic images for multi-person part segmentation. SYSU-Clothes~\cite{liang2016clothes} parses clothes via joint image segmentation and labels. BSANet~\cite{zhao2019multi} proposes to address object part parsing in the less explored multi-class setting, and designs a unified network architecture to solve this important problem. Grapy-ML~\cite{he2020grapy} proposes a novel graph pyramid module, which enables incorporating the hierarchical structural prior explicitly into feature learning via self-attention-based graph reasoning and progressive feature refinement. PRM~\cite{zhang2022human} is designed to generate features with adaptive context for various sizes and shapes of human parts.
HSSN~\cite{li2022deep} aims at a structured, pixel-wise description of visual observation in terms of a class hierarchy.
HIPN~\cite{liu2021hierarchical} proposes a new semi-supervised human parsing method for which the method only needs a small number of labels for training.
ISNet~\cite{jin2021isnet} proposes to augment the pixel representations by aggregating the image-level and semantic-level contextual information, respectively.
AFLA~\cite{liu2018cross} proposes a novel and efficient cross-domain human parsing model to bridge the cross-domain differences in terms of visual appearance and environment conditions.

Instance-level human parsing~\cite{Gong2018instance} is an interesting task, which aims to parse multiple human instances in a single pass. There are already several interesting works in this direction, including PNG~\cite{Gong2018instance}, NAN~\cite{zhao2020fine}, Holistic~\cite{li2017holistic}, Parsing R-CNN~\cite{yang2019parsing}, AIParsing~\cite{zhang2022aiparsing},  M-CE2P~\cite{Liu2019CE2P}.

Video human parsing~\cite{zhou2018adaptive,liu2014fashion} is also another related task, which parses every human body in the video data, which can be regarded as integrating video segmentation and instance-level human parsing. ATEN~\cite{zhou2018adaptive} first leverages convolutional gated recurrent units
to encode temporal feature-level changes, and the optical flow of non-key frames is wrapped with the temporal memory to generate their features. The following works are TimeCycle~\cite{wang2019learning}, UVC~\cite{li2019joint}, CRW~\cite{jabri2020space}, CLTC~\cite{jeon2021mining}, LIIR~\cite{li2022locality} and so on.

Some milestone human parsing methods are illustrated in Fig.~\ref{fig:timeline}. Given the large number of works
developed in the last few years, we only show some of the
most representative ones.

\subsection{Deep Learning-based Semantic Segmentation}
We review some of the most prominent deep learning-based semantic segmentation methods. In the human parsing community, many works use a semantic segmentation model as the image encoder of the methods, and re-train their model from those initial weights. This way accelerates the convergence of the network. According to their architectures, we divide into four groupings, e.g., fully convolutional networks, encoder-decoder based models, multi-scale networks, dilated convolutional models and deeplab family. It is noted that there are some parts shared by many methods, such as the encoder and decoder process, skip connections, and dilated convolution.

\begin{figure}[t]
 \begin{center}
 \includegraphics[width=7.5cm]{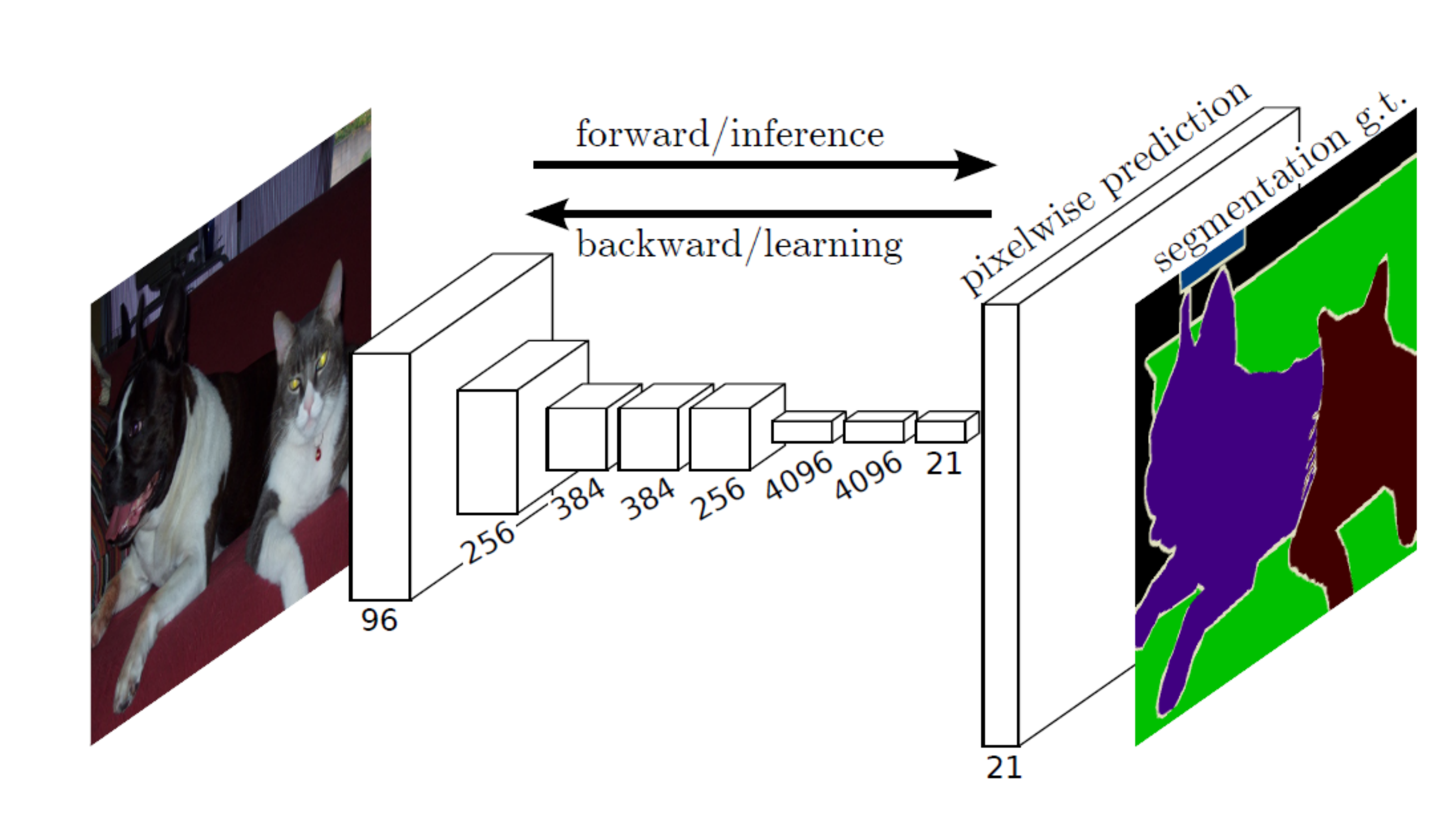}
 \caption{Fully convolutional networks, which can learn dense, per-pixel dense predictions for semantic segmentation. From~\cite{long2015fully}. }
 \label{fig:fcn1}
 \end{center}
\end{figure}

\textbf{Fully Convolutional Networks.}
Fully convolutional networks (FCN)~\cite{long2015fully} observes that the fully connected layers lose the spatial information and removes them from popular architectures (e.g., AlexNet~\cite{krizhevsky2017imagenet}, VGG~\cite{simonyan2014very}, GoogLeNet~\cite{szegedy2015going}). In this way, FCN includes only convolutional layers, which enables it to be applied to an image of any resolution, as shown in Fig.~\ref{fig:fcn1}. The method results in homochronous state-of-the-art results in several image segmentation datasets and is considered one of the most influential works in the area.

\begin{figure}[h]
 \begin{center}
 \includegraphics[width=8.8cm]{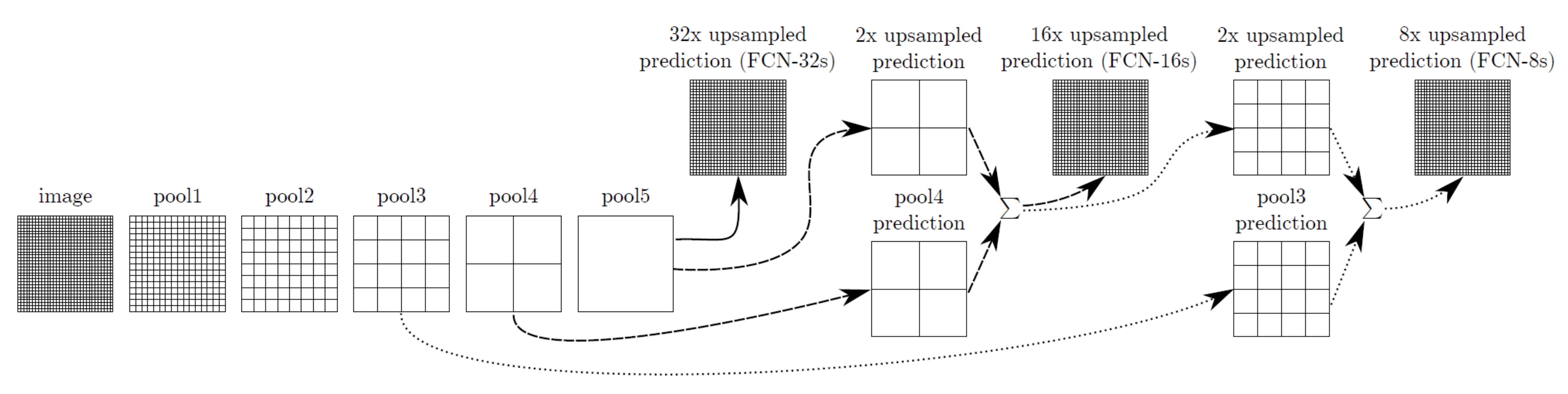}
 \caption{Skip connections combines semantics of deep layers and details of shallow ones to obtain accurate predictions. From~\cite{long2015fully}.}
 \label{fig:fcn2}
 \end{center}
\end{figure}

Skip connections link the outputs of nonadjacent layers by summing or concatenating, which combines the semantics of deep layers and details of shallow ones to obtain accurate segmentation results. FCN adds links that combine the final prediction layer with lower layers with finer strides (Fig.~\ref{fig:fcn2}). The most widely used architectures 'FCN-32s', 'FCN16s', and 'FCN8s' are obtained by the skip connection at different layers.
More dense skip connections for the same architecture are proposed for various applications~\cite{zhong2016fully}.

Although FCN can be trained in an end-to-end manner on any size images, it is not suited to deployment onboard mobile or other portable platforms due to its large amount of computation. To address this misalignment with more compact and efficient models, ENet~\cite{paszke2016enet} is proposed, one of the earliest real-time semantic segmentation. Comprising a larger encoder and very simple decoder, ENet is built out of several variations of bottleneck residual blocks comprising a dimensionality reduction, convolution, and provides similar or better accuracy with fewer parameters.

FCNs are considered revolutionary in many aspects. Since their high-performance and easy of deployment, they promote advances in medicine (e.g., brain tumor segmentation~\cite{wang2017automatic,minaee2021image}), driverless driving (e.g., ERFNet~\cite{holder2022efficient}) and so on.

\begin{figure}[t]
 \begin{center}
 \includegraphics[width=7.8cm]{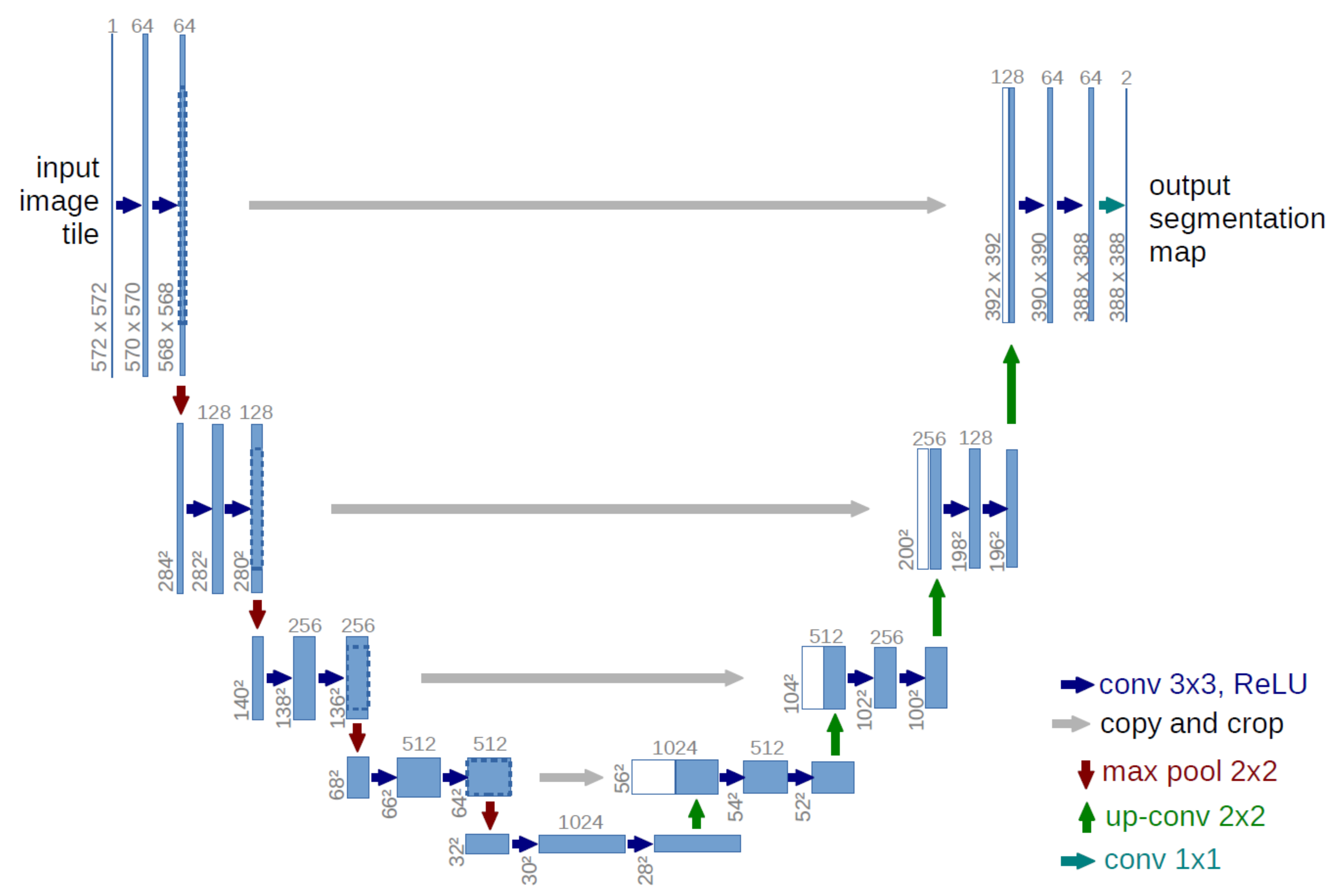}
 \caption{U-net model (an example for $572\times572$ resolution of input image). From~\cite{ronneberger2015u}.}
 \label{fig:unet}
 \end{center}
\end{figure}

\begin{figure}[h]
 \begin{center}
 \includegraphics[width=8.8cm]{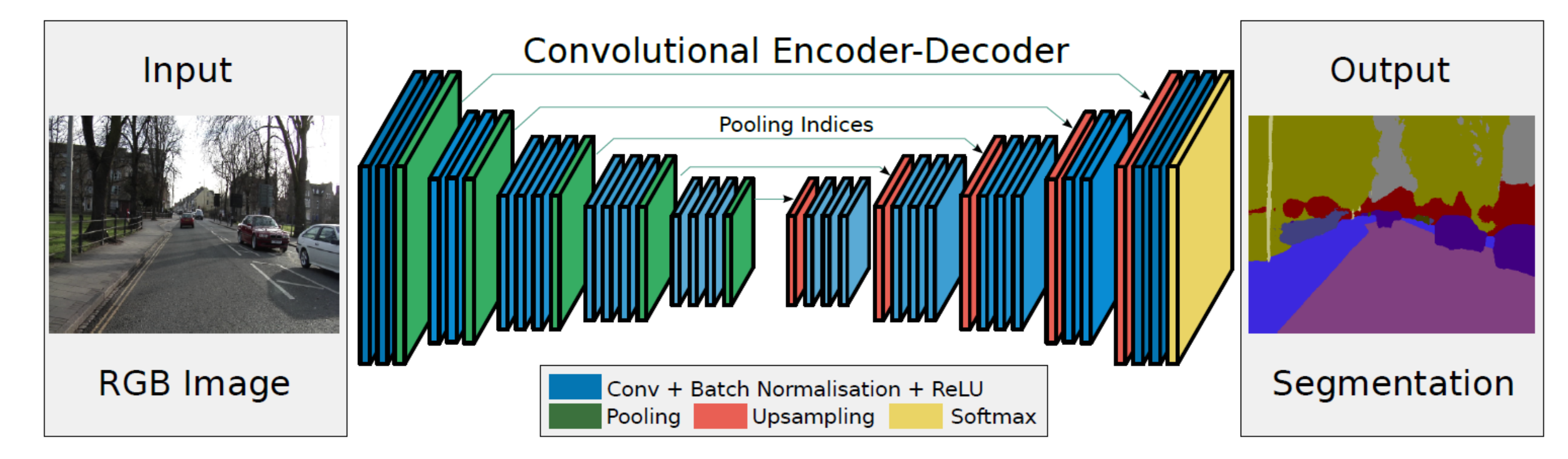}
 \caption{The SegNet architecture. The decoder upsamples its input using the pool indices from the encoder. From~\cite{Badrinarayanan2017SegNet}. }
 \label{fig:segnet}
 \end{center}
\end{figure}

\textbf{Encoder-Decoder Based Models.}
Hierarchical features created by pooling layers of FCN can partially lose some localization. Thus, we discuss other popular encoder-decoder based models which provide finer predictions.

The encoder-decoder models (as known as the U-nets) consist of two components (i.e., the encoder network and the decoder network). The encoder gradually reduces the spatial dimension by pooling or convolution layers, while the decoder recovers the resolution by upsampling. The decoder uses pooling indices computed in the max-pooling step of the corresponding encoder to perform upsampling, and concatenates corresponding features of the decoder. U-Net~\cite{ronneberger2015u} (Fig.~\ref{fig:unet}) and SegNet~\cite{Badrinarayanan2017SegNet} (Fig.~\ref{fig:segnet}) are well-known encoder-decoder algorithms. In this architecture, networks obtain rich details and semantics by the skip connections between the encoder and the decoder.

\begin{figure}[t]
 \begin{center}
 \includegraphics[width=8.8cm]{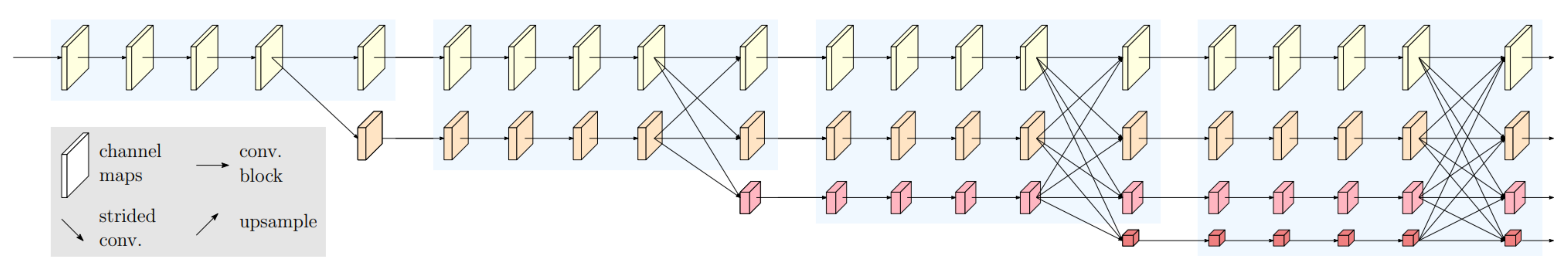}
 \caption{A simple example of a high-resolution network. The main difference between HRNetV1 and HRNetV2 lies in the outputs of the final stages. HRNetV1 only generates the high-resolution features, while HRNetV2 outputs all resolution features. From~\cite{sun2019high}.}
 \label{fig:hrnet}
 \end{center}
\end{figure}

Another popular model in this category is a high-resolution network (HRNets)~\cite{sun2019deep,sun2019high}, as shown in Fig.~\ref{fig:hrnet}. Different from U-Net and SegNet, they first reduce the spatial dimension and then recover it. HRNet maintains high-resolution representations through the encoding process by connecting the high-to-low resolution convolution streams in parallel, and repeatedly exchanging the information across resolutions. Many recent works on semantic segmentation, detection and facial landmark detection use HRNet as the backbone by exploiting contextual models.

\begin{figure}[h]
 \begin{center}
 \includegraphics[width=7.8cm]{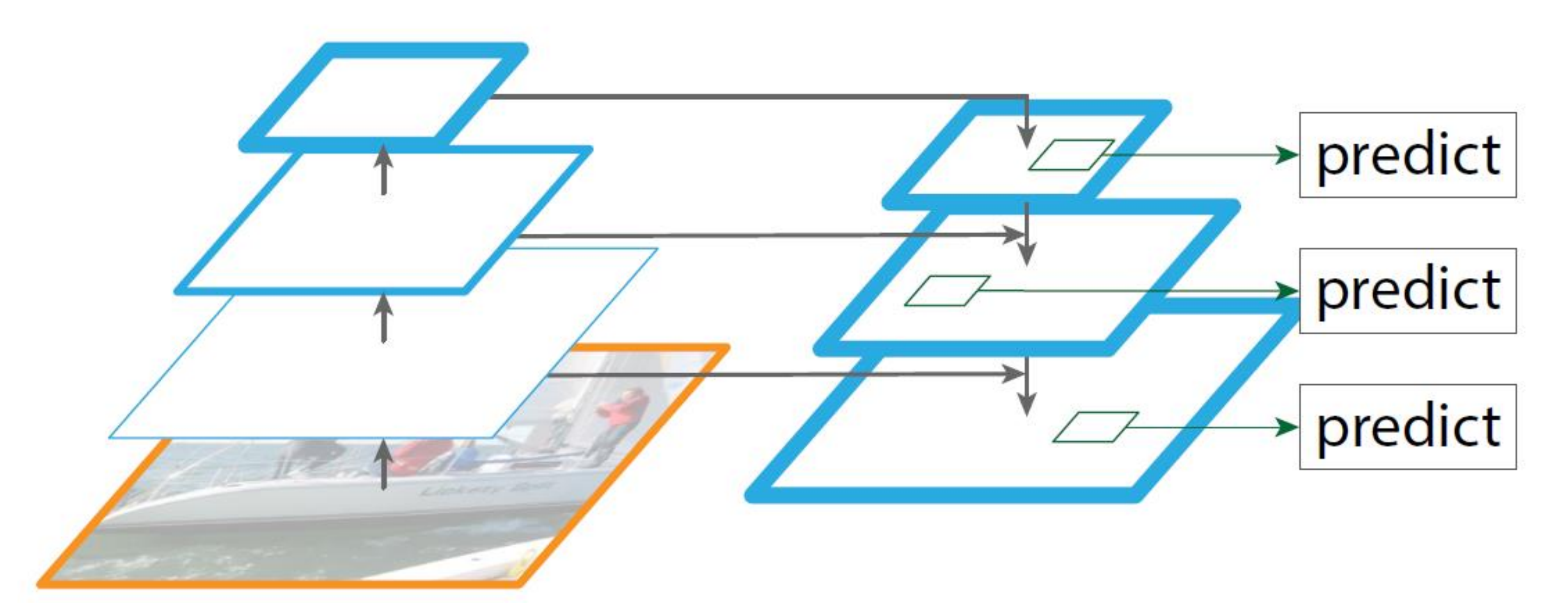}
 \caption{FPN involves a bottom-up pathway, a top-down pathway, and lateral connections. From~\cite{lin2017feature}. }
 \label{fig:fpn}
 \end{center}
\end{figure}

\textbf{Multi-Scale and Pyramid Network Based Networks.}
Multi-scale is an old method in image processing, which is employed in many computer vision tasks. One of the most prominent models of this sort is the Feature Pyramid Network (FPN)~\cite{lin2017feature} (Fig.~\ref{fig:fpn}), which is developed mainly for object detection and also applied to segmentation~\cite{zheng2021rethinking,minaee2021image}. Many methods use complex network structures to capture the inherent multi-scale context. FPN designs a simple and effective method that consists of a bottom-up pathway, a top-down pathway, and lateral connections. The concatenated feature maps are sent into a $3\times3$ convolution to generate the output of each stage. Finally, each stage of the top-down pathway produces a prediction to detect an object.

\begin{figure}[t]
 \begin{center}
 \includegraphics[width=8.8cm]{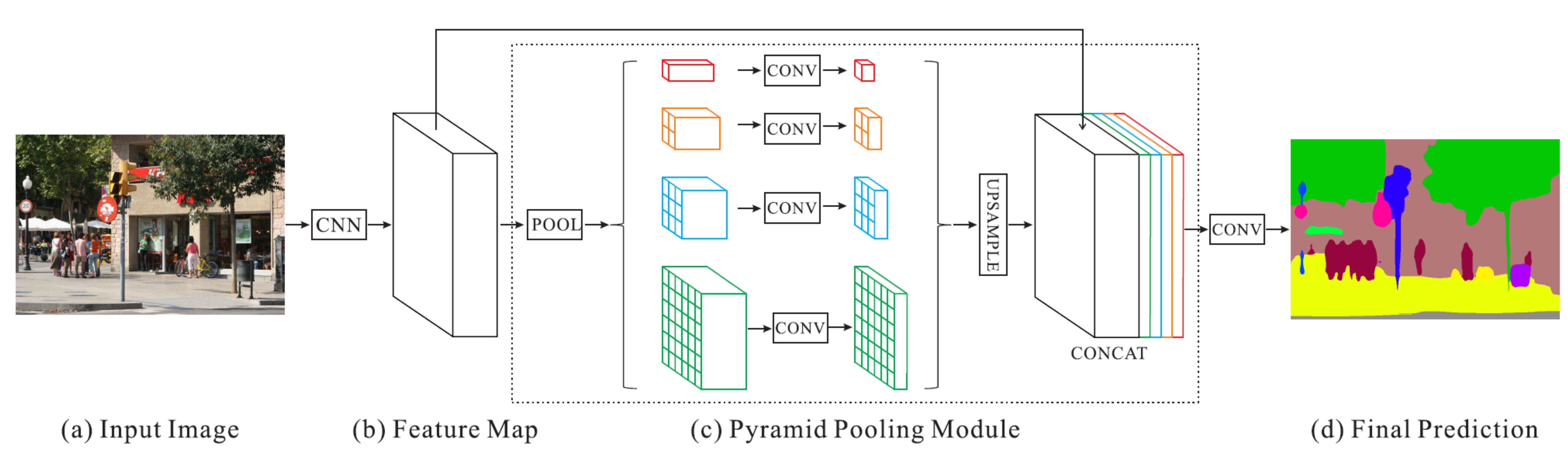}
 \caption{The overview of the PSPNet. The pyramid pooling module extracts different sub-regions by adopting varying-size pooling kernels in different strides. Upsampling and concatenation are used to form the final pixel-level prediction. From~\cite{Zhao2017Pyramid}. }
 \label{fig:pspnet}
 \end{center}
\end{figure}

PSPNet~\cite{Zhao2017Pyramid} uses the pyramid pooling module to capture the multi-scale global contexts for scene parsing, as shown in Fig.~\ref{fig:pspnet}. Given an input image, PSPNet first uses CNN to extract features, and then sends these features to the pyramid pooling model to harvest different scale pattern representations. There are four different scales, each scale corresponds to a pyramid level and is followed by a $1\times1$ convolutional layer to reduce dimensions. The outputs of the pyramid pooling module are upsampling and concatenating with the original features, which can obtain local and global context information. Finally, the representation is fed into a convolutional layer to generate the final pixel-level prediction.

Other models are using multi-scale analysis for semantic segmentation, such as LRR (Laplacian Pyramid Reconstruction)~\cite{ghiasi2016laplacian} designs a Laplacian pyramid for semantic segmentation, RefineNet~\cite{lin2017refinenet} proposes a multi-path refinement network for accurate prediction, and Multi-Scale Context Intertwining (MSCI)~\cite{lin2018multi} employs multi-scale context for segmentation.

\begin{figure}[h]
 \begin{center}
 \includegraphics[width=7.8cm]{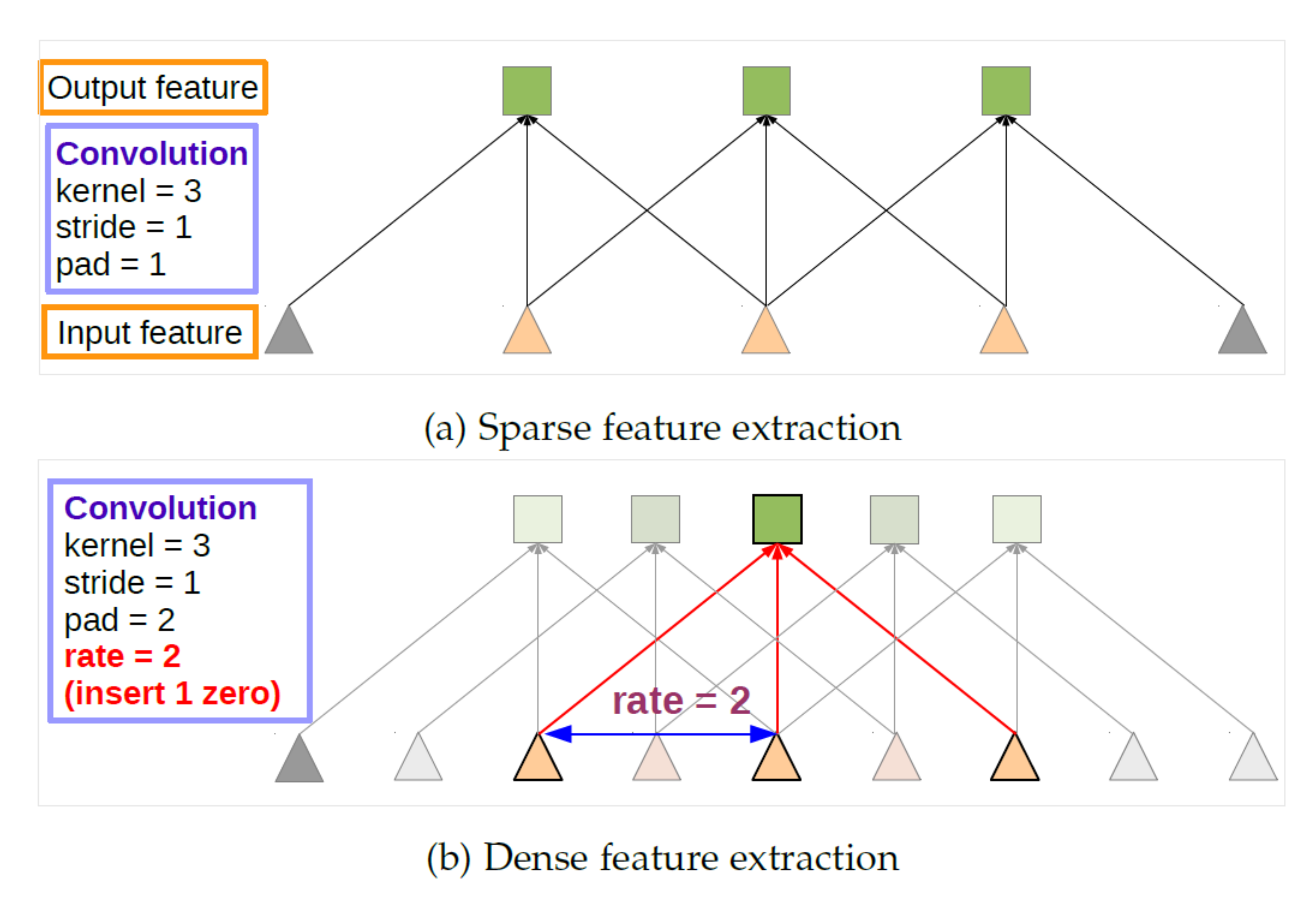}
 \caption{An example of dilated convolution. (a) Sparse feature extraction with standard convolution. (b) Dense feature extraction with dilated convolution with rate 2. From~\cite{Chen2017DeepLab}. }
 \label{fig:dilated convolution}
 \end{center}
\end{figure}
\textbf{Dilated Convolutional Models and DeepLab Family.}
The stack pooling and striding at consecutive layers reduces the resolution of the features, typically by a factor of 32 across each direction in recent DCNNs, affecting the segmentation accuracy of small objects. DeepLabv1~\cite{chen2014semantic} introduces atrous convolution (a.k.a. dilated convolution) which computes the responses of any layer at any desired resolution. Comparing the common convolution, atrous adds an extra parameter, rate parameter $r$, which corresponds to the stride with the input signal. It is defined as:
\begin{equation}
\label{eq:mm}
\begin{split}
&y[i] = x[i+r\cdot k]w[k],\\
\end{split}
\end{equation}
where the output $y[i]$ of atrous convolution of a 1-d input signal $x[i]$ with a filter $w[k]$ of length $k$. See Fig.~\ref{fig:dilated convolution} for illustration, for example, a $3\times3$ kernel with a dilation rate of 2 will have the same size of receptive field as a $5\times5$ kernel while using only 9 parameters, thus enlarging the receptive field with a negligible increase in computational cost.

\begin{figure}[h]
 \begin{center}
 \includegraphics[width=9.5cm]{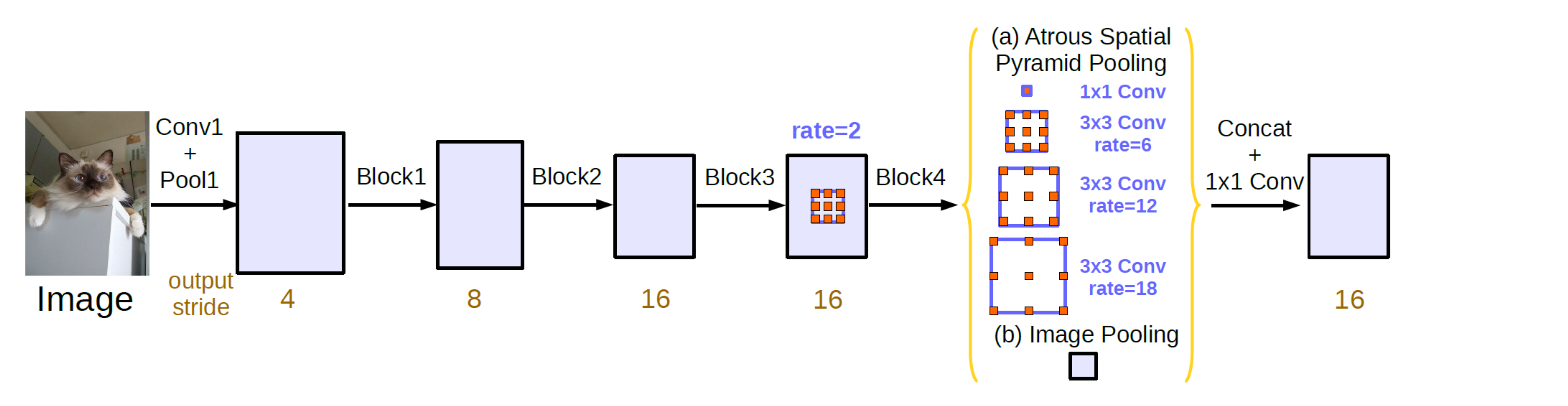}
 \caption{The DeepLabv3 model with image-level features. From~\cite{chen2017rethinking}. }
 \label{fig:DeepLabv3}
 \end{center}
\end{figure}

The DeepLab series is based on the FCN concept. Four versions of the deeplab series are called DeepLabv1~\cite{chen2014semantic}, DeepLabv2~\cite{Chen2017DeepLab}, DeepLabv3~\cite{chen2017rethinking} and DeepLabv3+~\cite{chen2018encoder}. DeepLabv1 lays the foundation for the other approaches. DeepLabv1 has two contributions. First, it uses atrous convolution to address the decreasing resolution in the network and increases the receptive field without adding computation. Second, it combines the responses at the final dcnn layer with a fully connected conditional random field (crf), which deals with the poor localization property of deep networks. DeepLabv2 proposes atrous spatial pyramid pooling (Aspp) to robustly segment objects at multiple scales. Aspp probes an incoming convolutional feature layer with filters at multiple sampling rates and effective fields-of-views, thus capturing objects and context at multiple scales.

Subsequently, DeepLabv3~\cite{chen2017rethinking} combines cascaded and parallel modules of atrous convolutions. The parallel convolution modules are grouped in the Aspp with image-level features encoding global context and further boost performance, as shown in Fig.~\ref{fig:DeepLabv3}. All the outputs are concatenated and processed by a $1\times1$ convolution to create the final prediction.

DeepLabv3+ employs an encoder-decoder architecture. It uses DeepLabv3 as the encoder and modifies the xception backbone with more layers. What's more, dilated depthwise separable convolutions instead of max pooling and batch normalization. The decoder first takes the low-level features from the network backbone as inputs and then concatenates the outputs of the Aspp. The simple yet effective decoder module refines the segmentation results along object boundaries.

\section{Human Parsing Datasets and Evaluation Metrics}
During the last decades, significant efforts have been made to develop various methods for human parsing. It is important to introduce some publicly available benchmark datasets and evaluation metrics. We also provide the quantitative performance of the promising models on popular datasets.

\begin{figure}[t]
 \begin{center}
 \includegraphics[width=9.0cm]{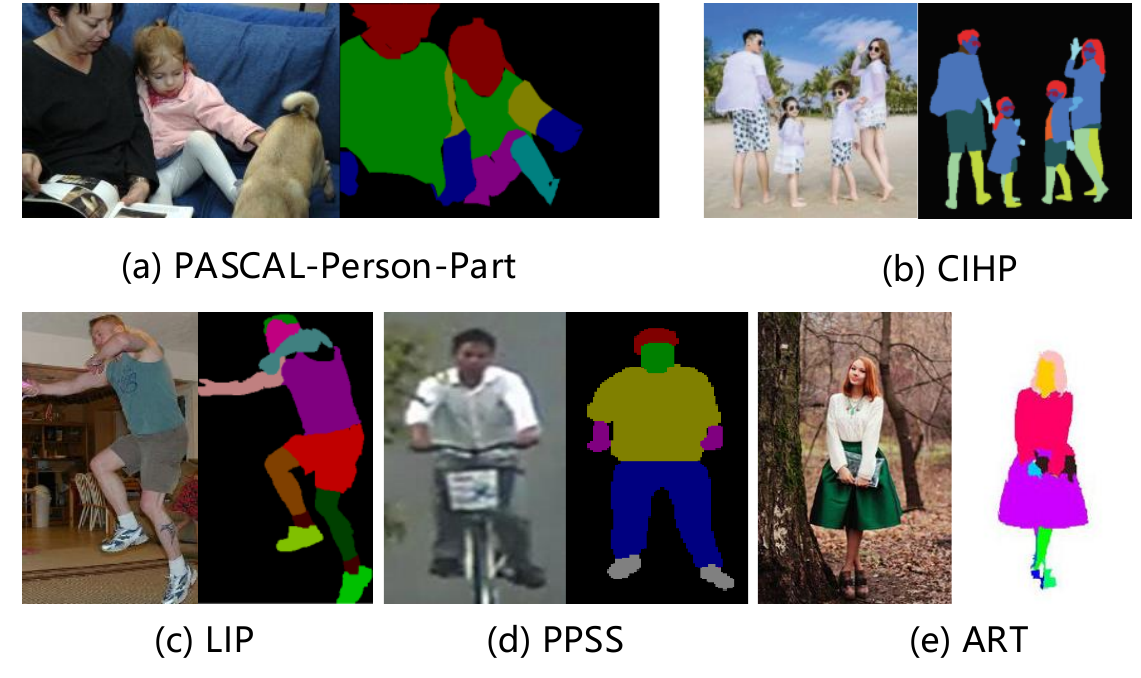}
 \caption{Five sample images with their corresponding ground truth from the PASCAL-Person-Part, LIP, CIHP, PPSS, and ATR, respectively. From~\cite{Chen2014Detect,gong2017look,Gong2018instance,luo2013pedestrian,Liang2015Deep}.}
 \label{fig:dataset}
 \end{center}
\end{figure}

\subsection{Datasets}

PASCAL-Person-Part~\cite{Chen2014Detect} is one of the most representative and widely used datasets in the human parsing task. There are multiple people appearances in an unconstrained environment. Its images mainly come from PASCAL VOC ~\cite{Chen2014Detect} dataset of image semantic segmentation.
This human parsing dataset has 7 classes: background, head, torso, upper-arm, lower-arm, upper-leg and lower-leg. This dataset is divided into two sets, training, and test, with 1716 and 1817 images, respectively. Fig.~\ref{fig:dataset} shows an example image and its pixel-wise label.

LIP~\cite{gong2017look} is one of the most popular datasets in the task, which is a single-person dataset. Its images are captured from a wider range of viewpoints, occlusions, and complex backgrounds. LIP defines 19 human parts (clothes) labels, including hat, hair, sunglasses, upper-clothes, dress, coat,
socks, pants, gloves, scarf, skirt, jumpsuits, face, right-arm, left-arm, right-leg, left-leg, right-shoe and left-shoe, and a background class. There are 50,462 images, including 30,362 for training, 10,000 for testing and 10,000 for validation. The testing set annotation is not disclosed, and the precision of the model on the testing set needs to be obtained through online evaluation.

CIHP~\cite{Gong2018instance} is a new large-scale multi-person benchmark with pixel-wise annotations on 19 semantic part labels, each image includes about three people. The images are collected from real-world scenarios, containing persons appearing with challenging poses and viewpoints, heavy occlusions, and a wide range of resolutions. This benchmark is divided into three sets, 28,280 images for training, 5,000 images for validation, and 5,000 for testing. There is a private test set for the actual challenge.

PPSS~\cite{luo2013pedestrian} is collected from 171 surveillance videos, it can reflect shading and lighting changes in real scenes. There are eight categories of the dataset, including hair, face, top, bottom, arms, legs, and shoes. There are 3673 annotated images in his dataset, including 1781 training images and 1892 test images.

Fashionista~\cite{Yamaguchi2012Parsing} consists of 158,235 images collected from Chictopia.com, a social networking website for fashion bloggers. It selects 685 images with good visibility of the full body for training and evaluation and is divided into ten folders, where the 9 folders serve as the training set, and the remaining 1 folder serves as the test set. This dataset includes a total of 56 categories, of which 53 are common clothing categories (e.g., dresses, bags, jackets, skirts, boots, sweaters) and the remaining three are hair, skin, and background.

Colorful Fashion Parsing Data (CFPD)~\cite{liu2013fashion} is a dataset containing color attributes of clothes and part categories. This dataset has 13 colors and 23 categories of tags. To balance the labeling efficiency and accuracy, they first over-segment each image into about 400 patches. There are 2,682 photos to form the final CFPD, randomly selecting half of these images as the training set, and taking the other half as the test set.

Daily Photos dataset~\cite{Dong2014A} consists of 2,500 images, all the pixels in the images are thoroughly annotated with 18 types of labels. The training: testing ratio is 2:1.

ATR~\cite{Liang2015Deep} is a combined dataset from four major sources: 685 images from the Fashionista; 2,682 images from the CFPD; 2,500 images from the daily photos and 1,833 images from the Human Parsing in the Wild (HPW). There are 18 categories in the dataset, including face, sunglasses, hat, scarf, hair, clothes, left arm, right arm, belt, pants, left leg, right leg, skirt, left shoe, right shoe, bag, dress, and other common 17 categories and 1 background.
This dataset includes 7,700 annotated images, including 6,000 images for training, 700 images for validation, and 1,000 for testing.

Fashion Clothing dataset~\cite{luo2018trusted} contains a total of 4,371 images from Clothing Co-Parsing (CCP)~\cite{yang2014clothing}, Fashionista~\cite{Yamaguchi2012Parsing} and Colorful Fashion Parsing Data (CFPD)~\cite{liu2013fashion}. It focuses more on human clothing details, and it consists of a background and 17 labels representing jewelry, bag, coat, suit, dress, glass, hair, pants, shoes, shirt, skin, skirt, upper-clothes, vest, and underwear, respectively. This
dataset can relatively reflect the network's capability to parse object details.

\subsection{Metrics for Segmentation Models}
We introduce the most popular metrics for assessing the accuracy of human parsing methods.

\textbf{Pixel Accuracy} is the ratio between the number of pixels correctly classified in the index dataset and the total number of pixels. The calculation formula is as follows:
\begin{equation}
	\label{eq:score}
    PA = \frac{{\sum\limits_i^C {{t_{ii}}} }}{{\sum\limits_i^C {\sum\limits_j^C {{t_{ij}}} } }},
\end{equation}
where $C$ denotes the number of categories, $t_{ii}$ denotes pixel $i$ is accurately classified as $i$, $t_{ij}$ denotes the real labels for $i$ pixel discriminant for $j$ class. The pixel accuracy ranges from 0 to 1, the higher the value, the better.

\textbf{Mean Pixel Accuracy (MPA)} is the extended version of PA, which averages PA over all classes, obtaining the final value.

\textbf{Intersection over Union (IoU)} or the \textbf{Jaccard Index} is one of the most commonly used metrics in human parsing. It calculates the ratio of the intersection and union between the predicted region of the class and the region of the real label class in each category. For the $k$ category, IoU is defined as follows:
\begin{equation}
	\label{eq:score}
     IoU = \frac{A\cap B}{A\cup B},
\end{equation}
where $A$ and $B$ denote the ground truth and the predicted segmentation maps, respectively.

\textbf{mean Intersection over Union (mIoU)} is the extended version of IoU, which averages over the total number of classes. It is widely used in reporting the performance of modern human parsing algorithms.

\textbf{F-1 score (F-1)} is a common metric used to evaluate the accuracy of a model. It combines precision and recall to measure uniformly and is calculated by the harmonic average of both indicators.
\begin{equation}
	\label{eq:score}
     F-1 = 2\cdot\frac{Precision\cdot Recall}{Precision + Recall}.
\end{equation}

\textbf{Foreground Pixel Accuracy (FGAcc)} only calculates the pixel accuracy of foreground human parts.

\subsection{Quantitative Performance of Methods}
In this section, we outline the performance of several of the previous methods on popular segmentation benchmarks. Since some models are not reported their performance on standard datasets, it is hard to make a full comparison. The following tables summarize the performances of several of the prominent models on different datasets. Table \ref{t1} focuses on the PASCAL-Person-Part test set. Since the introduction of FCN-based models, the performance has been greatly improved. Table \ref{t2} focuses on the LIP validation. On this dataset, the latest model improved by 32.01\% in terms of mIoU over the original model. Table \ref{t3} focuses on the CIHP test set. This dataset is more challenging than the PASCAL-Person-Part dataset, both two datasets are multi-person. Table \ref{t4} focuses on the ATR test set. The latest model achieves 87.16 in terms of F-1. Finally, Table \ref{t5} summarizes the performance of several models for the Fashion Clothing dataset.

\begin{table}
\begin{center}
\caption{Quantitative results on PASCAL-Person-Part test in terms of mean pixel Intersection-over-Union (mIoU) (\%)} \label{t1}
\resizebox{0.48\textwidth} {!} {
\begin{threeparttable}
    \begin{tabular}{ c|c|c|c|c }
    \toprule
     Method&Year&Pub.&Backbone&mIoU  	\cr
    \midrule
    \midrule
    HAZA~\cite{Xia2015Zoom}  &2016  &ECCV &VGG16 &57.54	\cr
    LIP~\cite{gong2017look}  &2017  &CVPR &ResNet101	 &59.36	\cr
    MMAN~\cite{Luo2018Macro} &2018  &ECCV &ResNet101	 &59.91	\cr
    Graph LSTM~\cite{liang2016semantic1}    &2016  &ECCV  &VGG16	 &60.61	\cr
    SE LSTM~\cite{Liang2017Interpretable}   &2017  &CVPR  &VGG16	 &63.57   \cr
    Joint~\cite{Xia2017Joint} &2017  &CVPR   &ResNet101	 &64.39	\cr
    MuLA \cite{nie2018mutual} &2018  &ECCV   &VGG16	 &65.1	\cr
    PCNet \cite{zhu2018pcn}   &2018  &AAAI   &PSPNet101	 &65.90 \cr
    Holistic  \cite{li2017holistic} &2017  &BMVC   &ResNet101	 &66.3	\cr
    WSHP \cite{fang2018weakly}    &2018  &CVPR   &VGG16	 &67.60	\cr
    SPGNet \cite{cheng2019spgnet}    &2019  &ICCV   &ResNet-101	 &68.36	\cr
    PGN \cite{Gong2018instance}   &2018  &ECCV   &ResNet101 &68.40   \cr
    RefineNet \cite{lin2017refinenet} &2017  &CVPR  &ResNet101	 &68.6	\cr
    CNIF \cite{wang2019learning}    &2019  &ICCV  &ResNet101	 &70.76	\cr
    DTCF~\cite{liu2020hybrid} &2020  &ACMMM   &ResNet101	 &70.80	\cr
    Graphonomy \cite{ke2019graphonomy}  &2019  &CVPR   &DeepLab v3+	 &71.14   \cr
    DPC \cite{chen2018searching} &2018  &NeurIPS  &-	 &71.34	\cr
    SNT \cite{2020Learning} &2020  &ECCV   &ResNet101	 &71.59	\cr
    NPPNet~\cite{zeng2021neural} &2021  &ICCV   &ResNet101 &71.73	\cr
    CDCL \cite{lin2019cross} &2021  &TCSVT   &ResNet101	 &72.82	\cr
    PRHP \cite{wang2021hierarchical} &2021 &TPAMI   &ResNet101	 &72.82	\cr
    BGNet~\cite{zhang2020blended}  &2020  &ECCV   &ResNet101	 &73.12   \cr
    GWNet~\cite{zhang2022grammar} &2022  &TIP   &ResNet101	 &74.67	\cr
    \bottomrule
    \end{tabular}
\end{threeparttable}
}
\end{center}
\end{table}

\begin{table}
\begin{center}
\caption{Quantitative results on LIP val in terms of mean pixel Intersection-over-Union (mIoU) (\%)} \label{t2}
\resizebox{0.48\textwidth} {!} {
\begin{threeparttable}
    \begin{tabular}{ c|c|c|c|c }
    \toprule
     Method&Year&Pub.&Backbone&mIoU  	\cr
    \midrule
    \midrule
    FCN-8s \cite{long2015fully}  &2015  &CVPR &VGG16 &28.29	\cr
    DeepLabV2 \cite{Chen2017DeepLab}  &2017  &TPAMI &ResNet101	 &41.64	\cr
    Attention \cite{Chen_2016_CVPR} &2016  &CVPR &VGG16	 &42.92	\cr
    DeepLab-ASPP \cite{Chen2017DeepLab}    &2017  &TPAMI  &VGG16	 &44.03	\cr
    LIP  \cite{gong2017look}   &2017  &CVPR &ResNet101	 &44.73   \cr
    MMAN \cite{Luo2018Macro} &2018  &ECCV   &VGG16	 &46.81	\cr
    JPPNet \cite{liang2018look}   &2018  &TPAMI   &PSPNet101	 &51.37 \cr
    CE2P \cite{Liu2019CE2P} &2019  &AAAI   &ResNet101	 &53.10	\cr
    BraidNet \cite{liu2019braidnet}    &2019  &ACMMM   &PSPNet	 &54.4	\cr
    SNT \cite{2020Learning}   &2020  &ECCV   &ResNet101 &54.8   \cr
    CorrPM~\cite{9157207} &2020  &CVPR  &ResNet101	 &55.33	\cr
    SLRS~\cite{li2020self} &2020  &CVPR  &ResNet101	 &56.34	\cr
    BGNet~\cite{zhang2020blended}  &2020  &ECCV   &ResNet101	 &56.82   \cr
    GWNet~\cite{zhang2022grammar} &2022  &TIP   &ResNet101	 &57.26	\cr
    CNIF~\cite{wang2019learning} &2019  &ICCV   &ResNet101	 &57.74	\cr
    DTCF~\cite{liu2020hybrid} &2020  &ACMMM   &ResNet101	 &57.82	\cr
    NPPNet~\cite{zeng2021neural} &2021  &ICCV   &ResNet101 &58.56	\cr
    HHP~\cite{wang2020hierarchical} &2020 &CVPR &DeepLabV3  &59.25	\cr
    SCHP~\cite{li2020self} &2020 &TPAMI &ResNet101  &59.36	\cr
    CDGNet~\cite{liu2022cdgnet} &2022 &CVPR  &ResNet101  &60.30	\cr
    \bottomrule
    \end{tabular}
\end{threeparttable}
}
\end{center}
\end{table}

\begin{table}
\begin{center}
\caption{Quantitative results on CIHP test in terms of mean pixel Intersection-over-Union (mIoU) (\%)} \label{t3}
\resizebox{0.48\textwidth} {!} {
\begin{threeparttable}
    \begin{tabular}{ c|c|c|c|c }
    \toprule
     Method&Year&Pub.&Backbone&mIoU  	\cr
    \midrule
    \midrule
    PGN \cite{Gong2018instance}   &2018  &ECCV   &ResNet101  & 55.80	\cr
    Graphonomy \cite{ke2019graphonomy} &2019  &CVPR &DeepLab v3+ & 58.58	\cr
    M-CE2P \cite{Liu2019CE2P}   &2019  &AAAI   &ResNet101  & 59.50	\cr
    CorrPM~\cite{9157207} &2020  &CVPR  &ResNet101 & 60.18	\cr
    BraidNet \cite{liu2019braidnet}  &2019  &ACMMM   &PSPNet  & 60.62	\cr
    SNT \cite{2020Learning}   &2020  &ECCV   &ResNet101  & 60.87	\cr
    PCNet~\cite{zhang2020part}  &2020 &CVPR  &ResNet101  & 61.05	\cr
    CDGNet~\cite{liu2022cdgnet} &2022 &CVPR  &ResNet101  &65.56	\cr
    \bottomrule
    \end{tabular}
\end{threeparttable}
}
\end{center}
\end{table}

\begin{table}
\begin{center}
\caption{Quantitative results on ATR test in terms of Foreground Pixel Accuracy (FGAcc) and F-1 score (F-1)} \label{t4}
\resizebox{0.48\textwidth} {!} {
\begin{threeparttable}
    \begin{tabular}{ c|c|c|c|c|c }
    \toprule
     Method&Year&Pub.&Backbone&F.G.Acc&F-1  	\cr
    \midrule
    \midrule
    ATR~\cite{Chen_2016_CVPR} &2015  &TPAMI &- &71.04   &64.38	\cr
    Attention~\cite{Chen_2016_CVPR} &2016  &CVPR &VGG16 &85.71   & 77.23	\cr
    Co-CNN~\cite{liang2015human} &2015  &ICCV &VGG16 &83.57   & 80.14	\cr
    LG-LSTM~\cite{Liang2016Semantic} &2016  &CVPR &VGG16 &84.79   & 80.97	\cr
    TGPNet~\cite{luo2018trusted}  &2018 &ACMMM   &VGG16  &87.91 & 81.76	\cr
    CNIF~\cite{wang2019learning} &2019  &ICCV   &ResNet101	 &85.51	\cr
    CorrPM~\cite{9157207} &2020  &CVPR  &ResNet101 &90.40 &86.12	\cr
    HHP~\cite{wang2020hierarchical} &2020 &CVPR &DeepLabV3  &89.23 &87.25	\cr
    CDGNet~\cite{liu2022cdgnet} &2022 &CVPR  &ResNet101 &90.19 &87.16	\cr
    \bottomrule
    \end{tabular}
\end{threeparttable}
}
\end{center}
\end{table}

\begin{table}
\begin{center}
\caption{Quantitative results on Fashion Clothing test in terms of Foreground Pixel Accuracy (FGAcc) and F-1 score (F-1).} \label{t5}
\resizebox{0.48\textwidth} {!} {
\begin{threeparttable}
    \begin{tabular}{ c|c|c|c|c|c }
    \toprule
     Method&Year&Pub.&Backbone&F.G.Acc&F-1  	\cr
    \midrule
    \midrule
    DeepLabV2~\cite{Chen2017DeepLab}  &2018  &TPAMI &VGG16 &56.08   &37.09 \cr
    Attention~\cite{Chen_2016_CVPR} &2016  &CVPR &VGG16 &64.47   &48.68	\cr
    TGPNet~\cite{luo2018trusted}  &2018 &ACMMM   &VGG16  &66.37 &51.92	\cr
    CNIF \cite{wang2021hierarchical} &2021 &TPAMI   &ResNet101  &68.59 &58.12	\cr
    PRHP \cite{wang2021hierarchical} &2021 &TPAMI   &ResNet101 &70.57 &60.19	\cr
    \bottomrule
    \end{tabular}
\end{threeparttable}
}
\end{center}
\end{table}

\section{Opportunities and Application}
Human parsing has made remarkable progress thanks to deep learning. We will introduce some of the promising research directions for further advancing human parsing algorithms.

\subsection{Weakly-Supervised and Unsupervised Learning}
Although supervised human parsing methods have demonstrated impressive results, their performance highly depends on the quantity and quality of training data. These data are a labor-intensive process that spends a significant amount of time and money. Weakly and Semi supervised~\cite{zhou2018brief} are becoming very active research communities. These approaches focus on knowledge transfer methods that migrate features with semantic labels to areas where collecting labels is difficult. In this way, the annotation of data can be reduced and the application scope can be expanded. For example, Fang~\emph{et al}. introduce pose-guide transfer~\cite{fang2018weakly} for human parsing.
Unsupervised learning is another promising direction that is attracting much attraction in various fields. In the field of segmentation, it adaptively divides different semantic parts by semantic consistency without the need for annotated data. Zhang~\emph{et al}~\cite{zhang2022contrastive} and Liu~\emph{et al}~\cite{liu2021unsupervised} design unsupervised human parsing methods which focus on simple gestures and cannot handle complex gestures or missing parts. More accurate methods should be further discovered and applied earlier.

\subsection{Total-body Human Parsing}
Social communication is a key function of human motion~\cite{birdwhistell2010kinesics}. People communicate tremendous amounts of information with the subtlest movements~\cite{joo2018total}, such as a glance and a smile. However, there are no existing systems simultaneously containing human parsing, face parsing, and hand parsing~\cite{liang2014parsing,lin2019face} to fully understand the pixel-wise spatial attributes of human in the wild. In this task, the toughest challenge is the total-body human parsing datasets. Labeling a complete and accurate pixel-level dataset is a huge task. A simple method is first to generate some false face and hand labels by high accuracy face parsing and hand parsing methods, respectively. And then false labels and ground truth of human parsing form the labels of total-body. Therefore, the integrated dataset and total-body human parsing methods need to be developed.

\subsection{Real-World Open-Set Methods}
In real-world human parsing tasks, limited by various objective factors, it is difficult to collect training samples to exhaust all parts. Therefore, most methods cannot guarantee consistently good performance in practical scenarios. A more realistic scenario is open set recognition (OSR)~\cite{geng2020recent}, where incomplete knowledge of the world exists at training time, and unknown classes can be submitted to an algorithm during testing,
requiring the classifiers to not only accurately classify the seen classes, but also effectively deal with unseen ones. Recently, some open-set methods are proposed in semantic segmentation. For example, DMLNet~\cite{cen2021deep} detects both in-distribution and out-of-distribution objects with an incremental few-shot learning module to gradually incorporate those OOD objects into
its existing knowledge base. The human body has various poses and many clothing types, and it is hard to define them all. Thus, real-world open-set methods are necessary.

\subsection{Memory Efficient and Real-time Approaches}
Many applications, such as mobile terminals, are critical to the parsing models that have a small amount of memory and/or can run in near real-time. To reduce memory, this can be done either by using simpler models, or by using model compression techniques or even training a complex model and then using knowledge distillation techniques to compress it into a smaller, memory efficient
network that mimics the complex model. Real-time is associated with a small amount of memory which speeds up inference and allows no less than 25 frames per second to be processed, achieving real-time.

\subsection{Interpretable Deep Models}
The deep learning-based human parsing method is a series of modeling methods for computers to improve prediction or behavior based on data. The current decision support system based on machine learning is a great improvement over the traditional rule-based method. Despite the advantages of machine learning models, users often question their decisions due to their lack of interpretability. To improve the transparency of machine learning models, building trust between users and machine learning models, and reducing potential risks in applications, such as bias in models, it is very necessary to provide model explanations.

\subsection{Application Scenarios}
Human parsing has been widely used in many applications such as person re-identification, pose estimation, and dress people.

\textbf{Person Re-identification.}
Person re-identification~\cite{zhu2020identity} aims to associate the person images captured by different cameras from various viewpoints, which attracts increasing attention from both the academia and the industry. However, part occlusions and inaccurate person detection can significantly change the visual appearance of a person in images and greatly increase the difficulty of this retrieval problem. Some methods~\cite{kalayeh2018human,song2018mask,li2021person} inject extra semantics in terms of the part to achieve the part alignment at the pixel-level. In this way, human parsing provides semantic information to help re-identification models perceive the position of the appearance of the human body.

\textbf{Pose Estimation.}
Pose estimation and human parsing are two crucial yet challenging tasks for human body configuration in 2D monocular images, these two tasks are highly correlated and could provide beneficial information for each other. Human parsing can facilitate localizing body joints in difficult scenarios. For example, MuLA~\cite{nie2018mutual} can fast adapt parsing and pose models to provide more powerful representations by incorporating information from their counterparts, giving more robust and accurate pose results. Dense pose estimation aims at mapping all human pixels of an RGB image to the 3D surface of the human body~\cite{guler2018densepose}. The mainstream dense pose estimation methods explicitly integrate human parsing supervision, such as
DensePose R-CNN~\cite{guler2018densepose}, Parsing R-CNN~\cite{yang2019parsing}.

\textbf{Dress People.}
The 3D reconstruction and modeling of humans from images is a central problem in computer vision~\cite{bhatnagar2019multi}. However, many methods~\cite{alldieck2019learning,alldieck2018detailed,alldieck2018video} lack realism and control. Multi-Garment Network (MGN)~\cite{bhatnagar2019multi} first models the inferring human body and layered garments on top as separate meshes from images directly. Human parsing provides fine-grained clothing details for dress people. Besides, Adaptive Content Generating and Preserving Network (ACGPN)~\cite{yang2020towards} predicts the semantic layout of the reference image that will be changed after try-on, and then determines whether its image content needs to be generated or preserved according to the predicted semantic layout. Beside, Zhang~\emph{et al}. propose a Decompose-and-aggregate Network (DaNet)~\cite{zhang2020learning} that densely
build a bridge between 2D pixels and 3D vertexes to facilitate the learning of 3D reconstruction.

\section{Conclusions}

We have surveyed recent human parsing algorithms based on deep learning models which have achieved impressive performance, and grouped them into five categories including structure-driven architectures, graph-based networks, context-aware methods, LSTM-based methods, and combined auxiliary information approaches. Then, we review quantitative performance analyses of these models on some popular benchmarks, such as PASCAL-Person-Part, LIP, CIHP, ATR, and Fashion Clothing datasets. Finally, we introduce some of the promising research directions for further human parsing algorithms and the application scenarios of the human parsing task.

\bibliographystyle{IEEEtran}
\bibliography{egbib}

\end{document}